\documentclass[10pt,journal,compsoc]{IEEEtran}

\usepackage{textcomp}
\usepackage{stfloats}
\usepackage{url}
\usepackage{bbm}
\usepackage{verbatim}
\usepackage{graphicx}
\usepackage{stfloats}
\usepackage{makecell}
\usepackage{pifont} 
\usepackage{booktabs}
\usepackage{ragged2e}
\usepackage{amssymb}
\usepackage{amsmath}
\usepackage{amsfonts}
\usepackage{algorithmic}
\usepackage{subfig}
\usepackage{marvosym}
\usepackage{stmaryrd}
\usepackage{adjustbox}
\usepackage[normalem]{ulem}
\usepackage[table]{xcolor}
\usepackage[breaklinks=true,bookmarks=false,urlcolor=black,colorlinks,citecolor=green]{hyperref}

\usepackage{multirow}
\usepackage{booktabs}
\usepackage{floatrow}
\newfloatcommand{capbtabbox}{table}[][\FBwidth]

\newcommand{\hlr}[1]{{\color{red}{#1}}}
\newcommand{\hlb}[1]{{\color{blue}{#1}}}
\newcommand{\hlg}[1]{{\color{green}{#1}}}
\newcommand{\hly}[1]{{\color{yellow}{#1}}}

\ifCLASSOPTIONcompsoc
  \usepackage[nocompress]{cite}
\else
  \usepackage{cite}
\fi

\ifCLASSINFOpdf
  
\else
  
\fi

\begin{document}
%
\title{SimPB++: Simultaneously Detecting 2D and 3D Objects from Multiple Cameras}
%
%
%
%

\author{Yingqi~Tang$^*$, Zhaotie Meng$^*$, Erkang Cheng$^\dagger$, and Haibin Ling,~\IEEEmembership{Fellow,~IEEE}

\IEEEcompsocitemizethanks{

\IEEEcompsocthanksitem Yingqi Tang, Zhaotie Meng, and Erkang Cheng are with Nullmax. E-mail: \{tangyingqi, mengzhaotie, chengerkang\}@nullmax.ai.

\IEEEcompsocthanksitem Haibin Ling is with Westlake University. E-mail: linghaibin@westlake.edu.cn.

\IEEEcompsocthanksitem $^*$ Equal contributions. $^\dagger$ Corresponding author.

\IEEEcompsocthanksitem A preliminary version~\cite{SimPB} of this work has appeared in ECCV 2024.

}


}

%
%

\markboth{Submitted to IEEE Journal}%
{Shell \MakeLowercase{\textit{et al.}}: Bare Demo of IEEEtran.cls for Computer Society Journals}

\IEEEtitleabstractindextext{%
\begin{abstract}

\justifying 
Simultaneous perception of 2D objects in perspective view and 3D objects in Bird's Eye View (BEV) presents a fundamental challenge for multi-camera autonomous driving systems. Existing approaches typically employ a two-stage pipeline with separate detectors, where 2D detection results serve only as a one-time initialization cue for 3D detection. In this paper, we propose \textit{SimPB++}, which \textbf{Sim}ultaneously detects 2D objects in the \textbf{P}erspective view and 3D objects in the \textbf{B}EV space from multiple cameras. It applies a unified paradigm that integrates 2D and 3D object detection into a single end-to-end model. Specifically, SimPB++ introduces a hybrid decoder architecture, where each layer couples a multi-view 2D decoder and a 3D decoder to perform their respective tasks interactively. To enable deep interaction between the two tasks, we introduce two novel modules: a Dynamic Query Allocation module that adaptively assigns 2D queries to relevant 3D candidates, and an Adaptive Query Aggregation module that progressively refines 3D representations using multi-view 2D features. This establishes a cyclic 3D-2D-3D refinement mechanism, allowing bidirectional feature enhancement across views. For multi-view 2D detection, we employ Query-group Attention to facilitate intra-group communication among 2D queries. To further improve long-range perception, we design a Crop-and-Scale strategy that adaptively zooms into distant regions across camera views, and a Propagating Denoising strategy that unifies denoising for both 2D and 3D queries, supported by an auxiliary RoI detector branch for foreground enhancement. Moreover, SimPB++ supports training with mixed supervision, combining 2D-only and fully annotated data, substantially reducing reliance on expensive 3D labels. Extensive experiments demonstrate that SimPB++ achieves state-of-the-art performance on the nuScenes benchmark for both 2D and 3D tasks, and excels in long-range detection (up to 150m) on Argoverse2, highlighting its effectiveness and scalability. 
Our code is available at:~\url{https://github.com/nullmax-vision/SimPB}.

\end{abstract}

}

\maketitle

\IEEEdisplaynontitleabstractindextext

\IEEEpeerreviewmaketitle

\IEEEraisesectionheading{\section{Introduction}\label{sec:introduction}}

\IEEEPARstart{C}{amera}-based perception has become a fundamental component of modern autonomous driving systems due to its low cost and rich semantic representation. 
Recent years have witnessed remarkable progress in 2D object detection in perspective views~\cite{FasterRCNN,YoloV1,FCOS}, which provides strong visual understanding of traffic scenes. 
Building upon these advances, monocular 3D object detection methods attempt to directly infer 3D object properties from a single camera and predict results in the bird's eye view (BEV) space~\cite{Monocular3DOD,M3D-RPN}. 
However, relying on a single frontal camera often limits the spatial coverage and robustness of the perception system. 
To address this limitation, multi-view 3D object detection methods~\cite{FCOS3D,DETR3D,BEVFormer} have been widely adopted in recent autonomous driving systems, where several surrounding cameras jointly provide a 360-degree perception of the environment.

\begin{figure}[t!]
  \centering
   \includegraphics[width=0.98\textwidth, trim=0 50 0 50, clip]{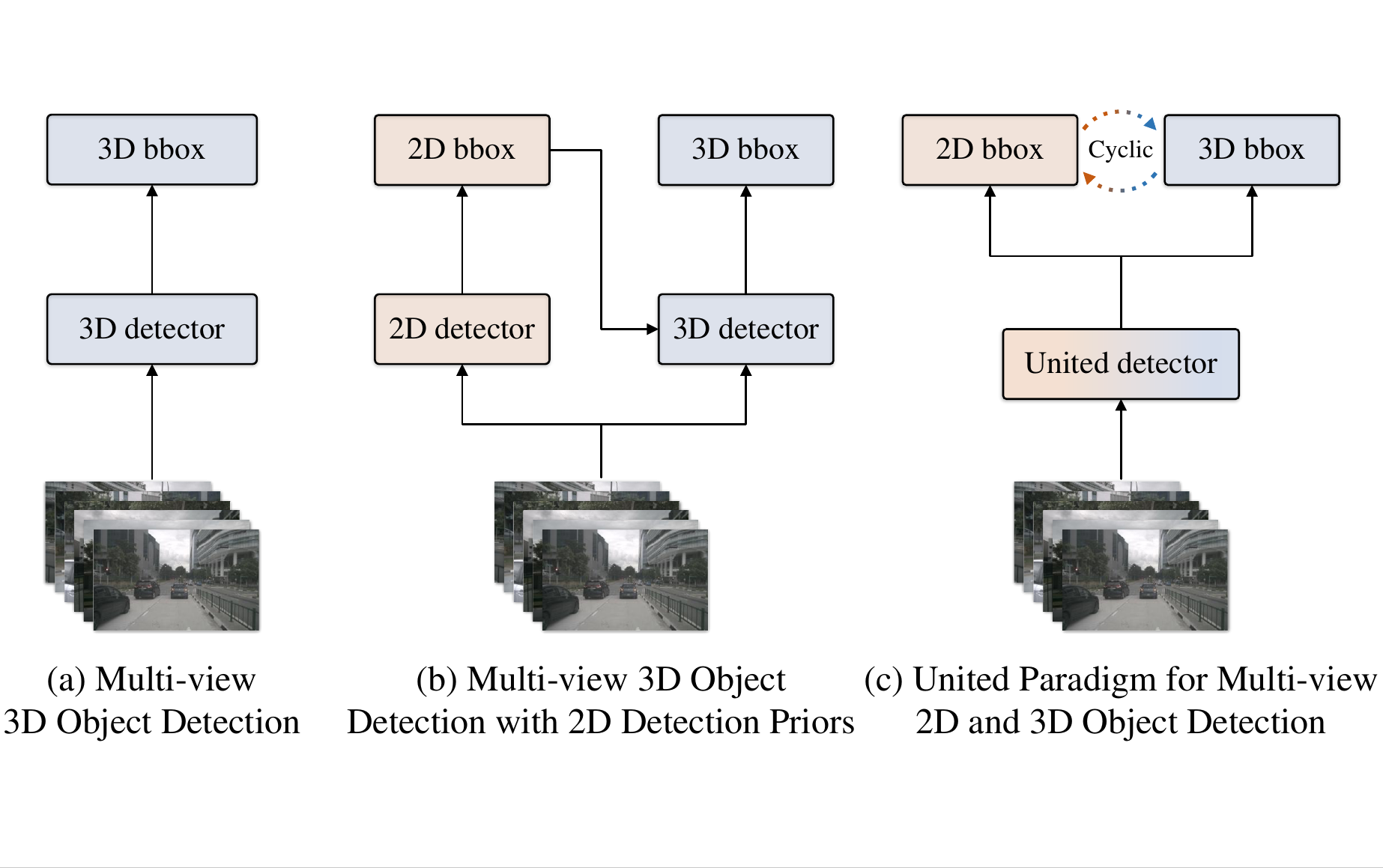}
  \caption{Comparisons of different multi-view object detection pipelines. (a) Multi-view 3D object detection. (b) A two-stage multi-view 3D object detector where 2D box detection is used as token selection or 3D query initialization. (c) Our proposed unified paradigm simultaneously predicts 2D and 3D results in a single model.}
  \vspace{-5pt}
  \label{fig:1_structure_compare}
\end{figure}

Existing multi-view 3D detection methods typically predict 3D objects directly from multi-camera images without performing explicit fusion of monocular predictions (Figure~\ref{fig:1_structure_compare} (a)). 
These methods can be broadly categorized into two paradigms: dense BEV-based approaches and sparse query-based approaches.
Dense BEV-based methods~\cite{LSS,BEVFormer,BEVDepth,BEVNeXt} first lift image features from multiple cameras into a unified BEV feature representation and then apply detection heads to estimate 3D objects. 
While this design provides a structured spatial representation, constructing high-resolution BEV features for long-range perception incurs substantial computational and memory costs. 
Alternatively, sparse query-based methods~\cite{DETR3D,PETR,Sparse4D,StreamPETR,SparseBEV} adopt a DETR-style detection framework~\cite{DETR} that directly interacts with multi-view image features using a set of learnable object queries, thereby avoiding dense BEV feature construction. 
These approaches often leverage backbones pre-trained on large-scale 2D detection tasks and have achieved competitive performance in camera-only 3D detection.

Despite these advances, 2D detection in perspective views remains an important source of information for multi-view perception. 
In practical systems, 2D detectors can provide reliable object localization cues that help identify potential 3D objects across different camera views. 
Consequently, several recent works~\cite{FocalPETR,MV2D,Far3D} attempt to improve 3D detection by incorporating 2D detection results from multiple cameras (Figure~\ref{fig:1_structure_compare}(b)). 
For example, MV2D~\cite{MV2D} converts 2D bounding boxes into 2.5D anchor points using ROI features and camera parameters, while Far3D~\cite{Far3D} lifts 2D boxes to 3D space with depth predictions from an auxiliary network. 
Focal-PETR~\cite{FocalPETR} further introduces instance-guided supervision to identify informative foreground tokens in perspective views. 
By leveraging strong 2D detection cues, these approaches improve 3D detection performance through query initialization or token selection.

However, the above two-detector paradigm still suffers from several fundamental limitations.
First, the 2D detector independently processes each camera view and treats the same physical object appearing in different images as separate instances. 
As a result, the subsequent 3D detector tends to focus on local observations from individual views rather than forming a holistic understanding of the object across cameras.
Second, the interaction between 2D and 3D detection is typically limited to a single stage, such as query initialization or foreground filtering, which prevents iterative refinement of object semantics between the two tasks.
Third, the use of heterogeneous architectures, e.g., CNN-based 2D detectors and Transformer-based 3D detectors, complicates model optimization and limits the potential benefits of joint training.

To address these limitations, we propose \textbf{SimPB++}, a unified query-based framework that \textbf{Sim}ultaneously performs 2D detection in \textbf{P}erspective views and 3D detection in \textbf{B}EV space using multiple cameras (Figure~\ref{fig:1_structure_compare}(c)). 
Our framework follows a DETR-style multi-view detection pipeline~\cite{DETR3D,Sparse4Dv3}, where multi-camera images are processed by a shared backbone and encoder to produce enriched image features. 
To jointly model 2D and 3D object representations, we design a hybrid decoder architecture that interleaves multi-view 2D decoding and 3D decoding within each decoder layer. 
Specifically, each hybrid decoder layer contains a 2D decoder module that refines perspective-view object queries and a 3D decoder module that updates BEV object queries.
To bridge the representation gap between 2D and 3D queries, we introduce a \emph{Dynamic Query Allocation} module that assigns 3D anchors to corresponding camera views based on geometric projections and forms grouped 2D queries for each image. 
These 2D queries capture view-specific object observations and are subsequently fused by an \emph{Adaptive Query Aggregation} module to produce updated 3D queries. 
Through this cyclic \textit{3D–2D–3D} interaction, SimPB++ enables continuous information exchange between perspective-view detection and BEV detection, allowing both tasks to mutually enhance each other during decoding.
Furthermore, we introduce a camera-aware query partitioning strategy that groups 2D queries according to camera indices. 
This design enables efficient query-group attention within each camera while preventing interference from irrelevant views.

To enhance the detection of distant targets, we propose a Crop-and-Scale strategy that generates additional camera views by cropping and scaling the original images, allowing the model to capture more fine-grained details of small and distant targets with dynamic feature sampling.
Specifically, the additional views create extra camera groups of 2D queries, improving the recall of 2D objects and aggregating multi-scale features from distant views in the 3D decoder layer.
We demonstrate that, at a modest computational cost, the strategy leads to significant improvements thanks to the sparse query detection framework with a fixed number of object queries.

Furthermore, to stabilize and accelerate convergence of the training process, we propose a unified denoising strategy called Propagating Denoising. 
This strategy constructs associated noise anchors in 2D and 3D from 3D ground truth, and corresponding noise queries are propagated between the 2D and 3D layers, facilitating improved feature alignment and enhancing the overall accuracy of the detection process.
We also enhance the foreground object feature by introducing an auxiliary branch without extra inference overhead.
This branch includes an RoI (Region of Interest) detector along with two depth estimators operating at instance and pixel levels. 
By incorporating these components, the model effectively focuses on the most relevant areas within an image while simultaneously deriving precise depth information.

The proposed joint 2D-3D framework allows SimPB++ to train with partially labeled data, enabling optimization using a combination of partially labeled (2D-only) and fully labeled (2D and 3D) annotations. Our results show that incorporating partially labeled (2D-only) annotations further enhances 3D object detection performance, thereby reducing the extensive costs associated with complete 3D labeling. 
In the experiments, SimPB++ is evaluated on the nuScenes dataset~\cite{Nuscene} and Argoverse2 dataset\cite{Argoverse2}. It achieves outstanding results on both 2D and 3D object detection on the nuScenes dataset, and long-range 3D object detection on the Argoverse2 dataset.

In summary, our major contributions are as follows:
\begin{itemize}
    \item We propose a novel unified query-based detector that detects simultaneously 2D objects in the perspective view and 3D objects in the BEV space from multiple cameras. 
    \item We introduce a hybrid decoder for 2D and 3D object inference. 
    A Dynamic Query Allocation module and an Adaptive Query Aggregation module are used to continuously update and refine the interaction between 2D and 3D results in a cyclic 3D-2D-3D scheme.  
    \item We design a Crop-and-Scale strategy to enhance distant target detections by leveraging a dynamic sampling and aggregation module, which provides more fine-grained details of the small and distant target.
    \item We introduce a Propagating Denoising strategy that leverages paired 2D and 3D noise queries for stable and accelerated convergence. Additionally, an Auxiliary Branch is utilized to enrich spatial and contextual information, which enhances overall detection accuracy.
    \item Our method supports partial-label training, effectively mitigating the prohibitive cost of 3D annotations for multi-view object detection.
    \item Our method achieves promising performance in both multi-view 3D object detection in the BEV space and 2D box detection in image space on the nuScenes and Argoverse2 dataset.
\end{itemize}

A preliminary version of this work, named SimPB, has been presented in ECCV 2024~\cite{SimPB}, which investigates the architecture of the hybrid 2D and 3D decoder layer in a single model by bridging 2D and 3D queries through a novel query allocation and aggregation mechanism.
In this work, we investigate the potential application of this joint 2D-3D paradigm and further enhance performance through novel strategies and architectural designs. Compared with~\cite{SimPB}, we bring the following new contributions: \textbf{(i)} We present a new Crop-and-Scale strategy for long-range detection. \textbf{(ii)} We develop a Propagating Denoising strategy to stabilize and accelerate convergence. \textbf{(iii)} We enhance detection accuracy by integrating an auxiliary branch that focuses on foreground objects. \textbf{(iv)} We explore the partial-labeling training, which potentially reduces the cost of 3D annotations. \textbf{(v)} We conduct extensive experiments on the additional Argoverse2 benchmark, with a specific focus on long-range target detection tasks. 

\section{Related Work}

\subsection{2D Object Detection}

Modern solutions to 2D object detection can be roughly divided into CNN-based methods~\cite{FasterRCNN, CascadeRCNN, LibraRCNN, YoloV1, CenterNet, FCOS} and Transformer-based approaches~\cite{DETR, DeformableDETR, ConditionalDETR, SparseDETR, DAB-DETR, DINO, CO-DETR}. CNN-based methods can be further categorized into two-stage and one-stage pipelines. Two-stage CNN models~\cite{FasterRCNN, CascadeRCNN, LibraRCNN} first create potential regional proposals and then refine them to an accurate bounding box. On the other hand, one-stage methods~\cite{YoloV1, CenterNet, FCOS} directly compute object information and greatly reduce the inference latency. 
More recently, Transformer-based method DETR \cite{DETR} formulates detection as a set prediction task and eliminates the need for complex post-processing steps, and is extended in many later works~\cite{ConditionalDETR, SparseDETR, DAB-DETR, DINO, CO-DETR} for improvements.
Despite the remarkable achievements in object detection using single-input images, the exploration of 2D object detection from multiple cameras remains relatively limited. 
In this paper, we focus on multi-view 2D object detection, where the association of 2D objects observed in different cameras is established by utilizing their corresponding 3D information in the BEV space.

\subsection{Multi-view 3D Object Detection}

Multi-view 3D object detection can be categorized into two main approaches: dense bird’s-eye view (BEV) methods~\cite{LSS, BEVDet, BEVDet4D, BEVFormer, BEVFormerV2, BEVDepth, BEVStereo, BEVNeXt, SOLOFusion, VideoBEV, HoP, GeoBEV} and sparse query-based methods~\cite{PETR, PETRv2, FocalPETR, StreamPETR, Far3D, DETR4D, Sparse4D, Sparse4Dv2, Sparse4Dv3, SparseBEV, DynamicBEV}.

Dense BEV methods typically construct an explicit BEV feature map from multiple cameras, followed by 3D object detection. One example is LSS~\cite{LSS}, which builds intermediate BEV features through interpolation from multi-view image features with a depth estimation step.
The BEVDet series~\cite{BEVDet, BEVDet4D} follows a similar approach to LSS, constructing the BEV space and utilizing additional data augmentation techniques to enhance 3D detection performance.
An alternative method to construct a BEV feature map is through the use of attention layers or MLP operators. For instance, BEVFormer~\cite{BEVFormer, BEVFormerV2} generates the BEV map using learnable grid-shaped BEV queries and employs deformable attention operators. Some dense BEV-based approaches~\cite{BEVDepth, BEVStereo, BEVNeXt} also explore depth supervision for accurate depth estimation to boost object localization.
Furthermore, several works~\cite{SOLOFusion, VideoBEV, HoP} improve object detection performance by propagating long-term historical features and wrapping them to the current timestamp.

Sparse query-based methods directly predict 3D objects from image features using learnable object queries. 
A line of research uses sampled points to establish interactions with multi-view images through camera parameters. 
For example, DETR3D~\cite{DETR3D} adapts DETR~\cite{DETR} by generating a single 3D reference point from an object query using camera parameters to gather image features.
DETR4D~\cite{DETR4D} and the Sparse4D series~\cite{Sparse4D, Sparse4Dv2, Sparse4Dv3} utilize additional reference points from 3D objects to effectively aggregate multi-view image features.
SparseBEV~\cite{SparseBEV} aggregates multi-scale features with adaptive self-attention, allowing for adaptive receptive fields.
DynamicBEV~\cite{DynamicBEV} leverages K-means clustering and top-K Attention to aggregate local and distant features.
Additionally, another approach focuses on learning flexible mappings between queries and image features via attention modules.
PETR~\cite{PETR, PETRv2} employs global attention by incorporating a 3D position encoding that contains geometric information for the mapping.
StreamPETR~\cite{StreamPETR} extends PETR~\cite{PETR} to a long-sequence 3D detection framework by a query propagation mechanism. 
DualViewDistill~\cite{DualViewDistill} combines a dense BEV with a sparse query framework to distill generative features from a foundation model.

Unlike traditional multi-view detectors that are confined to 3D outputs, SimPB++ introduces a hybrid decoder that unifies 2D and 3D tasks. By leveraging anchor projection to facilitate 3D-to-2D association, the framework eliminates redundant object representations across views and achieves seamless synchronization between 2D perspective and 3D BEV predictions.

\subsection{2D Auxiliary for 3D Object Detection}

Numerous studies have explored the benefits of utilizing 2D detectors as auxiliary information to improve 3D performance. Most approaches perform a detection task on the perspective view and transform the results for 3D query initialization.
For instance, BEVFormer v2~\cite{BEVFormerV2} introduces perspective 3D output as an additional form of supervision. It combines perspective 3D proposals with learnable 3D queries in a two-stage manner to enhance the detection performance.
MV2D~\cite{MV2D} generates 2D detection results on the image view and then lifts these 2D boxes using an implicit sub-network to form 3D queries.
Instead, Far3D~\cite{Far3D} integrates depth estimation with 2D detection results to generate reliable 3D queries. It significantly enhances the perception range and improves the detection performance of distant objects.
Different from them, FocalPETR~\cite{FocalPETR} performs 2D detection on the image view to generate salient foreground feature tokens, which are subsequently used for image feature aggregation along with 3D queries.

Our SimPB++ differs from the above approaches in several key aspects. Firstly, while those methods integrate 2D detectors in a two-stage manner, SimPB++ is an end-to-end and one-stage approach that simultaneously produces both 2D and 3D detection results. 
Secondly, in those approaches, the interaction between 2D and subsequent 3D detectors typically occurs only once in a 2D-3D manner. 
In contrast, SimPB++ adopts a cyclic 3D-2D-3D approach to interact with both 2D and 3D detectors. This cyclic interaction allows for continuous updates and refinement of the association between 2D and 3D results within the decoder.

\begin{figure*}[t]
  \centering
  \includegraphics[width=1\linewidth]{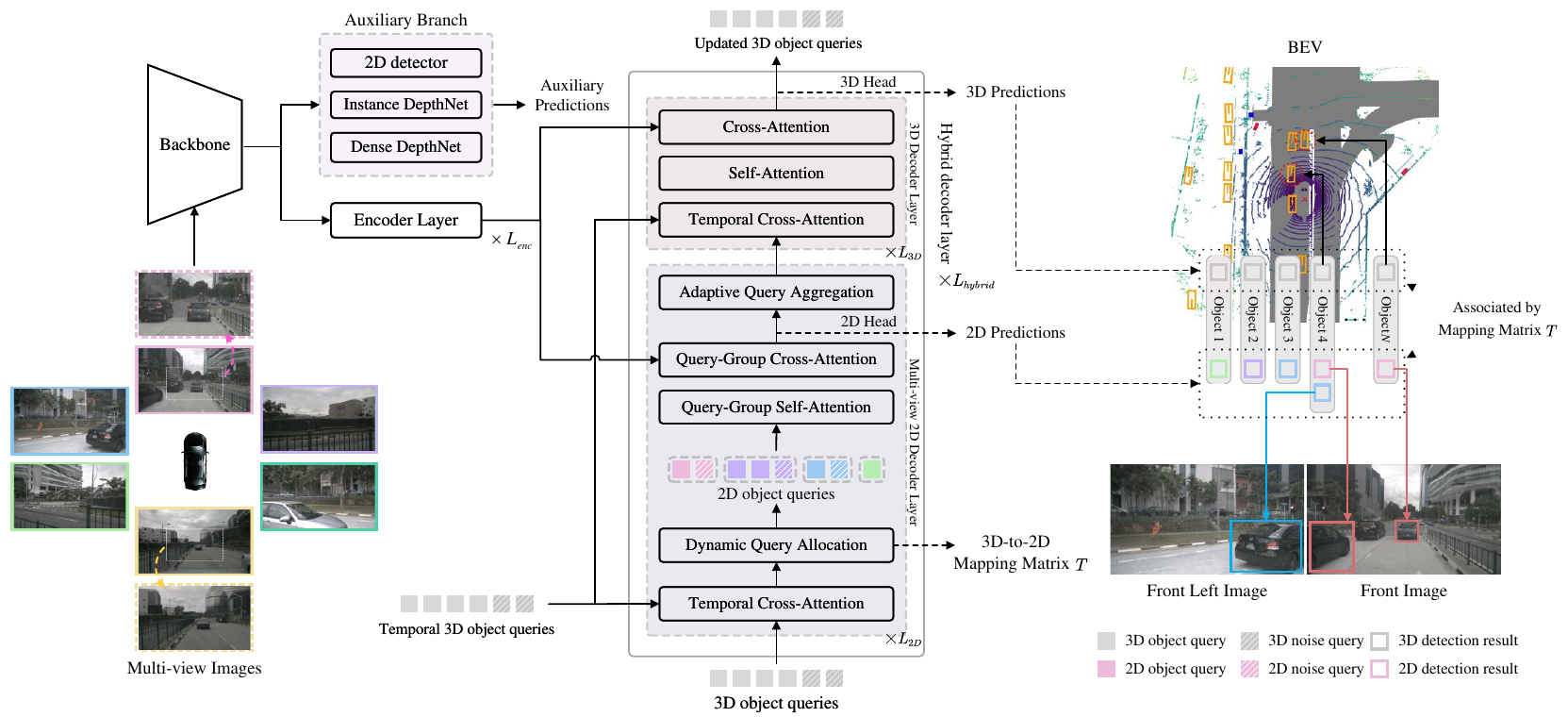}
  \caption{Overview of SimPB++, a unified multi-view 2D and 3D object detection framework. 
  An image backbone and an encoder module extract multi-view features.
  A hybrid decoder module, comprising multi-view 2D decoder layers and 3D decoder layers, is utilized to compute 2D and 3D detection results. An auxiliary branch is utilized to enhance feature representation in the training stage, while not used in inference.}
  \label{fig:2_framework_pipeline}
\end{figure*}

\subsection{Long-range 3D Object Detection}

Long-range 3D object detection remains a critical but challenging task in computer vision. Early monocular approaches estimate distance via object dimensions like pixel height and width~\cite{gokcce2015vision, haseeb2018disnet}, but their performance is highly sensitive to variations in object size and orientation. Subsequent methods attempt to directly regress distance~\cite{specific-distance}, yet these are constrained by their dependence on large, fully-annotated datasets. To mitigate the scarcity of 3D annotations for distant objects, recent works adapt more sophisticated strategies. For instance, R4D~\cite{li2022r4d} constructs a relational graph to infer distant targets using nearby objects as references, while LR3D~\cite{yang2024improving} learns a mapping from 2D boxes and depth to 3D parameters from close-range objects and generalizes it to long-range detection.
In the multi-view domain, methods often leverage 2D detectors and depth estimation. Far3D~\cite{Far3D}, for example, initializes 3D queries using features from regions of interest identified by a 2D detector and a depth network. Similarly, the approach in~\cite{khoche2024towards} employs dual detection networks for different ranges and augments LiDAR data with virtual points to enhance depth completion. SparseFusion~\cite{SparseFusion} addresses the challenge by introducing sparsity from both semantic and geometric perspectives, selectively lifting 2D image regions into the 3D space for efficient long-range inference.

Different from existing approaches, our proposed method develops a simultaneous 2D and 3D detection framework. Moreover, it introduces a Crop-and-Scale strategy with dynamic feature sampling tailored for high-resolution distant views. This mechanism captures the fine-grained details necessary for distant targets, enabling our method to outperform detectors specifically designed for long-range sensing.

\section{Method}

\subsection{Overview}

The overall architecture of our query-based SimPB++ is shown in Figure~\ref{fig:2_framework_pipeline}. Similar to the DETR-like framework~\cite{DETR, DETR3D, Sparse4Dv3}, SimPB++ consists of a backbone, an auxiliary branch, an encoder module with $L_{\text{enc}}$ layers, and a hybrid decoder module with $L_{\text{hybrid}}$ layers.
Given $V$ multi-view images $\{I_i\}_{i=1}^V $, multi-scale features $\{F_i\}_{i=1}^V$ are first extracted by a backbone (e.g. ResNet~\cite{ResNet} or V2-99~\cite{V2-99}) and followed by an encoder module to enhance expression. 
These multi-scale features are then fed into the hybrid decoder module for computing 2D and 3D object results.
Lastly, to fuse temporal information, we follow~\cite{StreamPETR} to propagate top-K 3D queries as temporal queries frame by frame.

\begin{figure*}[t]
  \centering
  \includegraphics[width=1\linewidth]{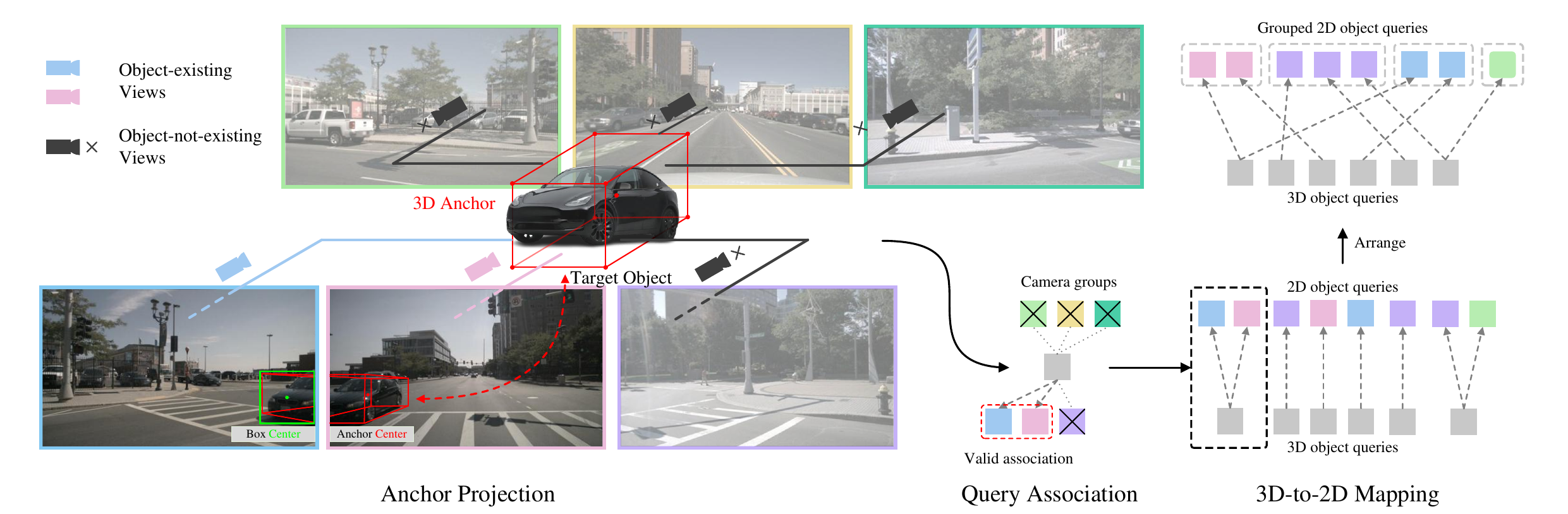}
  \caption{Dynamic Query Allocation. It establishes the query association between the BEV and the perspective view by applying 3D anchor projection with camera parameters.}
  \label{fig:3_allocation_module}
\end{figure*}

\subsection{Hybrid Decoder Module}

Standard transformer-based 2D detection approaches~\cite{DETR, DeformableDETR, DAB-DETR} apply a 2D decoder layer to extract 2D boxes from the input image. Similarly, popular transformer-based multi-view 3D detection methods~\cite{DETR3D, PETR, StreamPETR, Sparse4Dv3} employ a 3D decoder module to compute 3D objects in the BEV space. Different from them, we propose a hybrid decoder module that simultaneously calculates the 2D objects of each input camera and 3D objects in the BEV space. 
Furthermore, in SimPB++, the interaction between 2D and 3D results is continuously updated and refined in a cyclic 3D-2D-3D manner, which significantly enhances the overall 2D and 3D detection performance across multiple cameras.

Our proposed hybrid decoder module has $L_{\text{2d}}$ multi-view 2D decoder layers and $L_{\text{3d}}$ 3D decoder layers, each designed to compute 2D boxes in the perspective view and 3D object results in the BEV space, respectively.
The hybrid decoder module takes a set of $N$ 3D object queries, denoted as $Q_{\text{3d}} \in \mathbb{R}^{N \times C}$, along with their corresponding anchors $A_{\text{3d}} = \{(x, y, z, w, l, h, \theta, v_{x}, v_{y})\}$.
Within the multi-view 2D decoder layer, 3D object queries $Q_{\text{3d}}$ are passed through a Dynamic Query Allocation block, which allocates them to different camera groups to build 2D object queries $Q_{\text{2d}} \in \mathbb{R}^{M \times C}$.
Also, Query-group Self and Cross Attention are utilized to strengthen the interaction among 2D queries within each camera group and compute 2D results for each input camera. 
After obtaining 2D object detection results in the perspective view, these 2D queries are then processed by an Adaptive Query Aggregation block to reconstruct the 3D queries for the subsequent 3D decoder layers. 
The 3D decoder layer utilizes approaches in~\cite{PETR, DynamicBEV, StreamPETR} with self-attention and cross-attention operations and a 3D head to extract 3D objects.
Additionally, we incorporate a temporal cross-attention before both 2D and 3D layers to utilize the historical information by interacting with temporal queries.

\subsection{Dynamic Query Allocation}
\label{sec:allocation}

In the multi-view 2D decoder layer, a Dynamic Query Allocation block is designed to allocate 3D object queries to different camera groups and construct 2D object queries for multi-view 2D detection, as shown in Figure~\ref{fig:3_allocation_module}. 
A simple way is to uniformly distribute $N$ 3D queries to $V$ cameras and generate 2D queries. Instead, we utilize camera parameters to allocate a 3D query to its corresponding cameras and build related 2D queries. 

By interpreting 3D queries as 3D anchors, we can generate a set of $M$ valid 2D queries for $V$ input images. To establish the association between the 3D and 2D queries, we construct a 3D-to-2D mapping matrix $\mathbf{T} \in \mathbb{R}^{N \times M}$ based on the camera parameters.
In the mapping matrix $\mathbf{T}$, its element $\mathbf{T}(i,j) = 1$ indicates that the $i$-th 3D query is associated with the $j$-th 2D query. This matrix enables dynamic allocation and collection of 2D object queries:
\begin{equation}
    Q_{\text{2d}} = \mathbf{T}^\top  \cdot Q_{\text{3d}}.
\end{equation}

To determine the validity of a 3D query $q_n$, we begin by projecting its $K$ object points (e.g. center point and eight corner points of a 3D box) to $V$ image planes by camera intrinsic $\mathbf{I} \in \mathbb{R}^{3 \times 3} $ and extrinsic parameters $\mathbf{K} \in \mathbb{R}^{4 \times 4}$.
We obtain a set of points $P_v = \{p_1^v, p_2^v, \ldots, p_K^v \mid p_k^v=(u_k^v, v_k^v)\}$ on $v$-th image view, where $p_k^v$ is the projection of the $k$-th point. 
To determine if the 3D query $q_n$ is valid in the $v$-th image, we define a function $f(q_n, v)$ as:
\begin{equation}
f(q_n,v) = 
\begin{cases} 
    1 & \text{if}~\exists~p_k^v: 0<u_k^v<W, 0<v_k^v<H \\
    0 & \text{otherwise},
\end{cases}
\end{equation}
where $H$ and $W$ are the spatial resolution of an image. 

Considering $N$ 3D queries, each camera may have up to $N$ 2D query candidates. We initialize a diagonal matrix $\mathbf{T}_C^v \in \mathbb{R}^{N \times N}$ for the $v$-th camera, with the diagonal values set to 1. We retain only the valid 2D queries and eliminate the $j$-th column if $f(q_j, v) = 0$. We construct the mapping $\mathbf{T}^v \in \mathbb{R}^{N \times M_v}$ between the 3D queries and the $M_v$ valid 2D queries for the $v$-th camera, derived from $\mathbf{T}_C^v$. By concatenating $\{ \mathbf{T}^v \}_{v=1}^{V} $ along the column index, we obtain the mapping matrix $T$ between the $N$ 3D queries and the $M\sum_{v=1}^{V}M_v$ valid 2D queries.

In our implementation, $N$ is predefined for a fixed number 3D object queries, while $M$ is dynamically allocated depending on camera parameters and anchor projection.

\begin{figure*}[t]
  \centering
  \includegraphics[width=0.98\linewidth]{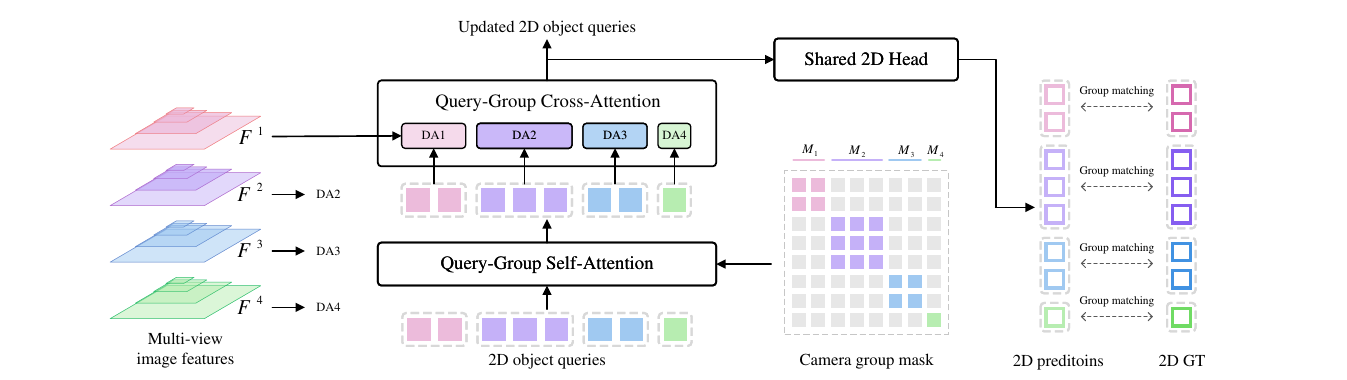}
  \caption{Query-Group Attention.
  We enforce interaction among 2D queries only within the same camera group.
  DA represents deformable attention.}
  \label{fig:4_query_group_attn}
\end{figure*}

\begin{figure*}[t]
  \centering
  \includegraphics[width=0.95\linewidth]{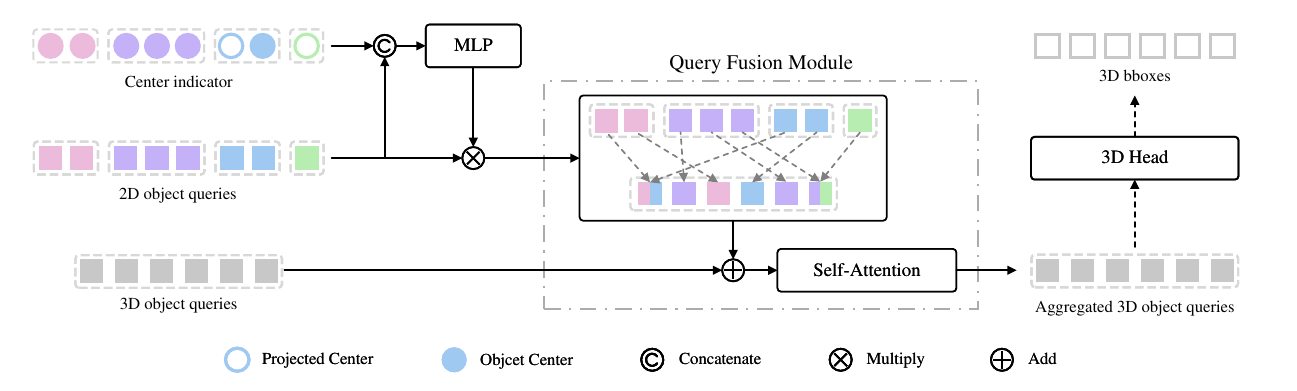}
  \caption{Adaptive Query Aggregation. The indicator vector represents whether a 2D query is truncated or not.}
  \label{fig:5_aggregation_module}
\end{figure*}

\subsection{Query Group Attention}

As described above, after dynamic query allocation, we have $M_v$ 2D queries for each camera $v$, resulting in a total of $M$ valid 2D queries.
To strengthen interactions within camera groups, we introduce query-group self-attention and query-group cross-attention.

Specifically, we first create a camera group mask $M_{\text{cg}} \in \mathbb{R}^{M \times M}$ filled with zeros. The value in the mask is then set to 1 if the queries originate from the same camera (as in Figure \ref{fig:4_query_group_attn}).
Additionally, we introduce an attention mask $\mathbf{\mathcal{M}} \in \mathbb{R}^{M \times M}$ as:
\begin{align}
\mathbf{\mathcal{M}}(i,j) = 
\left\{
\begin{array}{ll}
  0       & \text{if~} M_{\text{cg}}(i,j)=1\\
  -\infty & \text{otherwise}
\end{array}\right..
\end{align}
This attention mask ensures that queries within the same camera group can attend to each other while preventing queries from different camera groups from attending to one another. 

The query-group self-attention modifies the standard self-attention mechanism for the 2D query feature $\mathbf{X} \in \mathbb{R}^{M \times C}$ by incorporating an attention mask $\mathbf{\mathcal{M}}$:
\begin{align}
  \mathbf{X} = \text{softmax}\Big(\mathbf{\mathcal{M}} +\frac{\mathbf{Q}\mathbf{K}^{\text{T}}}{\sqrt{C}} \Big)\mathbf{V},
\end{align}
where $\mathbf{Q},~\mathbf{K}~\text{and}~\mathbf{V} \in \mathbb{R}^{M \times C}$ represent the 2D queries $Q_{\text{2d}}$ after linear transformations $W_Q$, $W_K$, and $W_V$, respectively. 

Next, query-group cross-attention is employed to facilitate effective information exchange, yielding precise 2D results for each independent input camera. 

In our multi-view 2D decoder, we use the projected object center as the reference point for a 2D box. In the case of object truncation, we rely on the center of the bounding rectangle of the projected anchor. 
Then, we utilize a shared 2D head across cameras to make predictions for the 2D bounding box and class.
Finally, these predictions are matched with the groundtruth in each groups through standard Hungarian matching algorithm as in DETR\cite{DETR}.

\subsection{Adaptive Query Aggregation}
\label{sec:aggregation}

After multi-view 2D detection, the 2D object queries are utilized to construct 3D queries for subsequent 3D object detection. As shown in Figure~\ref{fig:5_aggregation_module}, we propose an Adaptive Query Aggregation block to achieve the transformation by using the 3D-to-2D mapping matrix $\mathbf{T}$. 

To begin, we extend the 2D queries by incorporating information regarding the truncation status $\mathbbm{1}_{\text{center}} \in \{0, 1\}$ of a 2D object center: 
\begin{equation}
\tilde{Q}_{\text{2d}} = Q_{\text{2d}} \cdot \text{MLP}(\text{Concat}(Q_{\text{2d}}, \mathbbm{1}_{\text{center}})),
\end{equation}
where $\mathbbm{1}_{\text{center}} =0$ for a truncated 2D box where the box and anchor centers do not lie in the same image plane due to partial object visibility, as illustrated by the green box in Figure~\ref{fig:3_allocation_module}. Otherwise, we set $\mathbbm{1}_{\text{center}} =1$.
Consequently, it is essential to differentiate these truncated queries from non-truncated ones by incorporating distinct weight assignments. 

Next, to aggregate $\tilde{Q}_{\text{2d}} \in \mathbb{R}^{M \times C}$ into 3D queries, we replace the simple cross-attention with an integration of the 3D-to-2D mapping information:
\begin{equation}
Q_{\text{2d}}^{\text{fused}} = \frac{\mathbf{T} \cdot \tilde{Q}_{\text{2d}}}{\sum_{j=0}^{M}\mathbf{T}_j}.
\end{equation}
In this way, the 2D queries, which have been distributed from a shared 3D query during the Dynamic Query Allocation step, are aggregated back together.
Finally, we merge these aggregated queries with the original 3D queries using a residual connection and a self-attention operation: $Q_{\text{3d}}^{\text{agg}} = \text{Self-Attn}(Q_{\text{3d}} + Q_{\text{2d}}^{\text{fused}})$.
Additionally, we employ auxiliary 3D supervision to provide guidance for the aggregated 3D object queries.

\subsection{Crop-and-Scale Strategy}

Detecting distant targets is challenging because these objects appear small in the image view, particularly when using a downsampled low-resolution input for efficient model inference. This results in insufficient and indistinct features for small targets, often leading the model to sample adjacent background features. Consequently, a detection algorithm may struggle not only to distinguish long-range objects but also to accurately determine their positions and estimate their attributes.

\begin{figure}[t]
  \centering
  \includegraphics[width=0.86\linewidth]{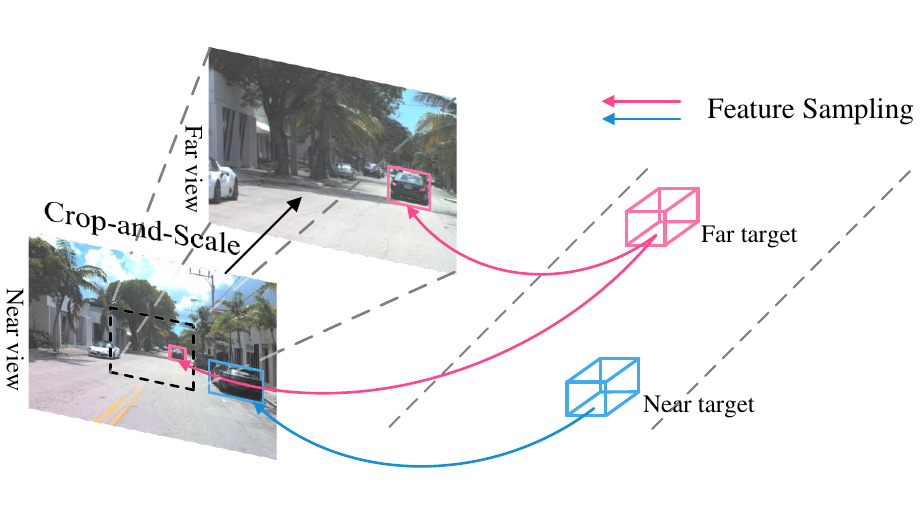}
  \caption{
  The Crop-and-Scale strategy generates a high-resolution distant view and employs a dynamic sampling mechanism to provide enhanced features for long-range object detection.
  }
  \label{fig:scale_crop_pipeline}
\end{figure}

To address this, we introduce a Crop-and-Scale strategy within our framework to generate new distant views that preserve fine-grained details, as illustrated in Figure~\ref{fig:scale_crop_pipeline}. Our observation is that distant targets tend to appear in consistent, view-specific regions. For example, in the front view, distant content is typically centered, whereas in the front-right view, it is often concentrated toward the left. Leveraging this regularity, we zoom into these regions to create additional long-range perspectives that capture most distant targets and augment them with richer detail.
Thanks to our dynamic sampling mechanism, long-range targets can sample and aggregate more detailed information from this newly created view, while near-range targets maintain the same features since their anchors are not projected onto this long-range view.

Instead of transforming model inputs $I\in \mathbb{R}^{H\times W\times 3}$, we apply cropping and scaling to the original images $I_o \in \mathbb{R}^{H_{o}\times W_{o}\times 3}$ to prevent aliasing.
Since model inputs are typically downsampled for computational efficiency, applying spatial transformations at such reduced resolutions would degrade image quality due to interpolation artifacts.

Specifically, given a defined crop region, we first extract the corresponding patch $I_c\in \mathbb{R}^{H_c\times W_c\times 3}$ from the original high-resolution image. This cropped segment is then rescaled to the model's input dimensions to produce the final long-range view $I_{cs}\in \mathbb{R}^{H\times W\times 3}$.
The crop regions are strategically set based on the localized clusters of distant objects within each specific viewpoint.
For instance, in the front and rear views, the crop is centered on the focal region; in the front-right view, the left edge of the crop is aligned with the scaled view's left boundary, while ensuring that the horizon aligns with the vertical coordinate of the focal point. The resulting image $I_{cs} \in \mathbb{R}^{H\times W\times 3} $ is then used as an additional distant camera view alongside the multi-view image inputs.

\subsection{Propagating Denoising}

Denoising~\cite{DN-DETR} is an effective technique originally developed for transformer-based methods such as DETR~\cite{DETR} to accelerate model convergence. It has since been widely adopted in transformer-based object detection for both 2D~\cite{DN-DETR,DAB-DETR,DINO} and 3D tasks~\cite{StreamPETR,RayDN}.
While numerous existing studies have applied this technique to either 2D or 3D detection, to the best of our knowledge, no prior work has explored a unified denoising formulation that operates jointly across both.
To improve the 2D-3D interaction, we introduce a propagating denoising mechanism that enables mutual transformation of noisy queries between 2D and 3D spaces. This design promotes joint convergence and strengthens the association between 2D and 3D predictions.

\begin{figure}[t]
  \centering
  \includegraphics[width=0.89\linewidth]{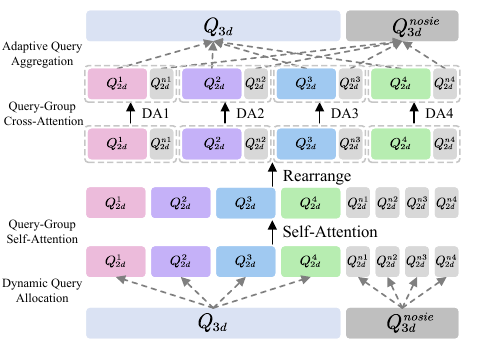}
  \caption{The Propagating Denoising in multi-view 2D decoder layers. DA represents Deformable Attention; we omit the temporal cross attention for clarity.}
  \label{fig:denoising_pipeline}
\end{figure}

Concretely, we first initialize 3D noisy queries $Q^{\text{noise}}_{\text{3d}}$ and construct noisy 3D anchors $A^{\text{noise}}_{\text{3d}}$ by adding noise to the ground-truth annotations. Unlike conventional detection methods that directly construct 2D noisy anchors from 2D ground truth, we instead generate noisy 2D queries and anchors via the proposed query allocation strategy. Specifically, we associate the 3D noisy queries and anchors with multiple camera views, yielding grouped 2D noisy queries $Q^{\text{noise}}_{\text{2d}}$ and corresponding grouped 2D noisy anchors. Importantly, this 3D-to-2D mapping is not performed by projecting the noisy 3D anchors; instead, we rely on ground-truth associations to ensure precise alignment.
We then reorganize the permutation of grouped 2D queries along with their associated noise queries and attention masks to facilitate grouped self-attention and cross-attention, as illustrated in Figure~\ref{fig:denoising_pipeline}. Finally, the updated grouped 2D noisy queries are aggregated to restore 3D queries for subsequent processing.

\subsection{Auxiliary Branch}

The original SimPB framework in our preliminary study employs a hybrid decoder for joint 2D and 3D detection. However, we observe that 2D information is not fully exploited during feature extraction. To address this limitation, we introduce an Auxiliary Branch to enhance feature representation, as illustrated in Figure~\ref{fig:2_framework_pipeline}. This module comprises three components: a 2D Region of Interest (RoI) detector, an instance DepthNet, and a dense DepthNet. The 2D RoI detector identifies foreground objects, while the instance DepthNet estimates their corresponding depths. Both predictions are supervised using the same 2D ground truth as the multi-view 2D decoder layers. Additionally, a lightweight dense DepthNet predicts a dense depth map, supervised by projected LiDAR point clouds. Unlike prior works~\cite{MV2D, Far3D} that use detected foreground objects and estimated depths to initialize coarse queries, we employ the auxiliary branch solely for supervision during training. It is discarded at inference time, thereby introducing no additional computational overhead.

\subsection{Loss Functions}
The training loss consists of both the multi-view 3D object detection loss and the 2D object detection loss, along with an auxiliary loss:
\begin{equation}
    \mathcal{L}_{\text{overall}} = \mathcal{L}_{\text{3d}} + \mathcal{L}_{\text{2d}} + \mathcal{L}_{\text{aux}}.
\end{equation}

We employ the same 3D detection loss  $\mathcal{L}_{\text{3d}}$ as described in \cite{Far3D, BEVNeXt, Sparse4Dv3}.
For the 2D detection loss, we adopt $\mathcal{L}_{\text{detr2d}}$ from DETR~\cite{DETR, DeformableDETR} and incorporate the observation angle~\cite{alpha_angle} loss of a 3D box in the image view.
\begin{equation}
\mathcal{L}_{\text{2d}} = \mathcal{L}_{\text{detr2d}} + \lambda_1 \mathcal{L}_{\text{alpha}}.
\end{equation}
As typically done, we encode \textit{alpha}-angle with \textit{sin} and \textit{cos} functions: 
\begin{equation}
    \mathcal{L}_{\text{alpha}} = \frac{1}{M} \sum_{i=1}^{M} |\sin{(\theta_i)} - \widehat{\sin}_{\theta_i}| + |\cos{(\theta_i)} - \widehat{\cos}_{\theta_i} |.
\end{equation}
where $\widehat{\sin}_{\theta_i}$ and $\widehat{\cos}_{\theta_i}$ represent the estimated sine and cosine components of the observation angle $\theta_i$ of an object.
The auxiliary loss comprises three components, as follows: 
\begin{equation}
    \mathcal{L}_{\text{aux}} = 
    \mathcal{L}_{\text{roi2d}} + 
    \mathcal{L}_{\text{ins\text{-}depth}} +
    \lambda_2 \mathcal{L}_{\text{dense\text{-}depth}}.
\end{equation}
It integrates RoI loss (including classification and regression loss) for foreground detection\cite{Far3D}, focal loss for multi-bin instance depth prediction, and L1 loss for dense map depth estimation. 
The balancing weights for the loss terms $\lambda_1, \lambda_2$ are empirically set as $0.5$ and $ 0.2$, respectively.

\begin{table*}
\centering

\caption{The implement details on nuScenes and Argoverse2 dataset.}
\resizebox{0.85\textwidth}{!}
{
\begin{tabular}{l|cc|ccccc}
    \toprule
     Dataset & Backbone & Resoluation & Epoch & Ablation Epoch & Batch size & Learning rate & weight decay \\
    \midrule
    \multirow{2}{*}{nuScenes\cite{Nuscene}}
    &  ResNet50  & $704\times 256$ & 100 & 24 & 6 & 4e-4 & 1e-3\\
    &  ResNet101 & $1408\times 512$& 100 & -  & 2 & 2e-4 & 1e-3\\
    \midrule
    Argoverse2\cite{Argoverse2}
    &  V2-99     & $960\times 640$ & 10  & 3  & 2 & 2e-4 & 1e-3\\
    \bottomrule
\end{tabular}
}
\label{tab:implement_details}
\end{table*}

\section{Experiments}
\subsection{Dataset and Metrics}

We evaluate our method using the nuScenes~\cite{Nuscene} and Argoverse2~\cite{Argoverse2} benchmarks.

The nuScenes~\cite{Nuscene} dataset is a multi-modal video dataset consisting of 1000 videos, which are split into training, validation, and testing sets, with 700, 150, and 150 videos, respectively. Each video is about 20s long, captured by 6 cameras and annotated with Lidar data at a frequency of 2Hz.
The dataset contains a total of 1.4M annotated 3D bounding boxes for 10 classes.
The ground truth of 2D bounding box is generated from 3D labels following methods in~\cite{MV2D, FocalPETR, StreamPETR}.
For 3D object detection, we employ the official evaluation metrics of nuScenes, including mAP, mATE, mASE, mAOE, mAVE, mAAE, and NDS.
Additionally, we evaluate the multi-view 2D object detection using box detection metrics following the standard COCO protocol~\cite{COCODataset} on the nuScenes validation set.

The Argoverse2~\cite{Argoverse2} dataset is particularly used to evaluate performance on detecting distant objects. It consists of 1000 scenes with data collected at a frequency of 10 Hz. It features 26 labeled categories. The dataset is officially partitioned into a training set of 700 scenes, a validation set of 150 scenes, and a testing set of 150 scenes. For evaluation, the Argoverse 2 metrics are recommended, which include mAP, CDS, ATE, ASE, and AOE metrics.

\subsection{Implementation Details}
Following previous works~\cite{Sparse4Dv3, Far3D}, we use ResNet50~\cite{ResNet}, ResNet101~\cite{ResNet}, and V2-99~\cite{V2-99} as backbones. 
To effectively leverage multi-view image features in large-scale scenarios, when using ResNet101 with an input resolution of $1408 \times 512$, we employ a single deformable transformer encoder layer and initialize 1200 queries. For all other configurations, we omit the encoder layer and use 900 queries to maintain efficiency while ensuring fair comparisons.
In the hybrid decoder, we set $L_{\text{2d}}=1$ and $L_{\text{3d}}=1$ in each hybrid layer $L_{\text{hybrid}}$ to include one multi-view 2D decoder layer and one 3D decoder layer. The hybrid decoder consists of 3 hybrid layers with a total of 6 sub-layers.
The scale-and-crop strategy is applied to the front and rear views. Consequently, 2 additional views are generated as inputs for nuScenes, while 3 additional views are generated for Argoverse2 because the rear view in Argoverse2 contains rear-left and rear-right views.
The experiments are trained on 8 NVIDIA A800 GPUs and tested on an NVIDIA 4090 GPU.
We do not employ Test Time Augmentation (TTA), CBGS\cite{CBGS}, or future frames during training.
The models are trained for 100 epochs on nuScenes \cite{Nuscene} and 10 epochs on Argoverse2 \cite{Argoverse2}, using the AdamW optimizer with a weight decay of 0.001. Other detailed training parameters under different settings are listed in Table \ref{tab:implement_details}.

\begin{table*}[t]
\centering
\caption{Comparison results of 3D detection on nuScenes validation dataset. * The backbone benefits from perspective pertaining. 
}
\resizebox{0.97\textwidth}{!}
{
\begin{tabular}{@{}l|c|c|cc|ccccc@{}}
    \toprule
    Method  & \ \, Backbone\ \,  & \ \, Resolution\ \,& \ mAP$\uparrow$\, & \, NDS$\uparrow$\  & \,mATE\,$\downarrow$ & \,mASE\,$\downarrow$ & \,mAOE\,$\downarrow$ & \,mAVE\,$\downarrow$ & \,mAAE\,$\downarrow$ \\

    \midrule
    SparseBEV*\cite{SparseBEV}   & ResNet50 &$704\times 256$& 0.448& 0.558& 0.595& 0.275& 0.385& 0.253&  \textbf{0.187} \\
    StreamPETR*\cite{StreamPETR} & ResNet50 &$704\times 256$& 0.450& 0.550& 0.613& 0.267& 0.413& 0.265& 0.196 \\
    BEVNext*\cite{BEVNeXt}       & ResNet50 &$704\times 256$& 0.456& 0.560& 0.530& 0.264& 0.424& 0.252& 0.206 \\
    DynamicBEV*\cite{DynamicBEV} & ResNet50 &$704\times 256$& 0.464& 0.570& 0.581& 0.271& 0.373& 0.247& 0.190 \\
    RayDN*\cite{RayDN}           & ResNet50 &$704\times 256$& 0.469& 0.563& 0.579& 0.264& 0.433& 0.256& 0.187 \\
    Sparse4Dv3\cite{Sparse4Dv3}     & ResNet50 &$704\times 256$& 0.469& 0.561& 0.553& 0.274& 0.476& 0.227& 0.200 \\
    CorrBEV*\cite{CorrBEV}       & ResNet50 &$704\times 256$& 0.475& 0.574& 0.598& 0.253& 0.335& 0.246& 0.200 \\
    GeoBEV*\cite{GeoBEV}         & ResNet50 &$704\times 256$& 0.479& 0.575& \textbf{0.496}& 0.261& 0.438& 0.236& 0.216 \\
    \rowcolor{gray!15}
    SimPB*                       & ResNet50 &$704\times 256$& 0.487& 0.590& 0.536& 0.261& 0.346& 0.208& \textbf{0.187} \\
    \rowcolor{gray!15}
    SimPB++*               & ResNet50 &$704\times 256$& \textbf{0.505}& \textbf{0.603}& 0.535& \textbf{0.258}& \textbf{0.305}& \textbf{0.197}& 0.193 \\
    
    
    \midrule
    SOLOFusion\cite{SOLOFusion}     & ResNet101 &$1408\times 512$& 0.483& 0.582& 0.503& 0.264& 0.381& 0.246& 0.207 \\
    BEVNext*\cite{BEVNeXt}       & ResNet101 &$1408\times 512$& 0.500& 0.597& 0.487& 0.260& 0.343& 0.245& 0.197 \\
    SparseBEV*\cite{SparseBEV}   & ResNet101 &$1408\times 512$& 0.501& 0.592& 0.562& 0.265& 0.321& 0.243& 0.195 \\
    StreamPETR*\cite{StreamPETR} & ResNet101 &$1408\times 512$& 0.504& 0.592& 0.569& 0.262& 0.315& 0.257& 0.199 \\
    Far3D*\cite{Far3D}           & ResNet101 &$1408\times 512$& 0.510& 0.594& 0.551& 0.258& 0.372& 0.238& 0.195 \\
    DynamicBEV*\cite{SparseBEV}  & ResNet101 &$1408\times 512$& 0.512& 0.605& 0.575& 0.270& 0.353& 0.236& 0.198 \\
    CorrBEV*\cite{CorrBEV}       & ResNet101 &$1408\times 512$& 0.517& 0.606& 0.554& \textbf{0.247}& 0.298& 0.241& \textbf{0.189} \\
    RayDN*\cite{RayDN}           & ResNet101 &$1408\times 512$& 0.518& 0.604& 0.541& 0.260& 0.315& 0.236& 0.200 \\
    GeoBEV*\cite{GeoBEV}         & ResNet101 &$1408\times 512$& 0.526& 0.615& 0.458& 0.254& 0.318& 0.238& 0.207 \\
    
    Sparse4Dv3*\cite{Sparse4Dv3} & ResNet101 &$1408\times 512$& 0.537& 0.623& 0.511& 0.255& 0.306& 0.194& 0.192 \\
    \rowcolor{gray!15}
    SimPB*                       & ResNet101 &$1408\times 512$& 0.539& 0.629& \textbf{0.475}& 0.260& 0.280& 0.192& 0.197 \\
    \rowcolor{gray!15}
    SimPB++*               & ResNet101 &$1408\times 512$& \textbf{0.549}& \textbf{0.639}& 0.490& 0.258& \textbf{0.233}& \textbf{0.187}& \textbf{0.189} \\
    
    \bottomrule
\end{tabular}
}
\label{tab:nus_val}
\end{table*}

\begin{table*}[t]
\centering
\caption{Comparison results of 2D detection on nuScenes val dataset. * The backbone benefits from perspective pretraining.}
\resizebox{0.8\textwidth}{!}{  
\begin{tabular}{@{}l|c|c|cccccc@{}}
    \toprule
    Method \ & \, Backbone \, & \, Resolution \, & \, AP \, & \, AP$_{50}$\, & \, AP$_{75}$\, & \, AP$_S$\, & \, AP$_M$ \, & \, AP$_L$ \, \\
    \midrule
    StreamPETR*\cite{StreamPETR}         & ResNet50 & $704\times256$ & 0.205 & 0.404 & 0.184 & 0.014 & 0.129 & 0.319 \\
    MV2D*\cite{MV2D}                     & ResNet50 & $704\times256$ & 0.226 & 0.456 & 0.198 & \textbf{0.054} & \textbf{0.196} & 0.297 \\
    DeformableDETR\cite{DeformableDETR}     & ResNet50 & $704\times256$ & 0.230 & 0.465 & 0.201 & 0.028 & 0.156 & 0.339 \\
    \rowcolor{gray!15}
    SimPB*                               & ResNet50 & $704\times256$ & 0.256& 0.495 & 0.237 & 0.044 & 0.177 & 0.361 \\
    \rowcolor{gray!15}
    SimPB++*                       & ResNet50 & $704\times256$ & \textbf{0.266} & \textbf{0.510}& \textbf{0.247}&0.045 & 0.185&\textbf{0.371} \\
    \midrule
    StreamPETR*\cite{StreamPETR}         & ResNet101 & $1408\times512$ & 0.249 & 0.465 & 0.240 & 0.042 & 0.191 & 0.344 \\
    MV2D*\cite{MV2D}                     & ResNet101 & $1408\times512$ & 0.271 & 0.523 & 0.250 & 0.047 & 0.204 & 0.367 \\
    DeformableDETR\cite{DeformableDETR}     & ResNet101 & $1408\times512$ & 0.250 & 0.502 & 0.222 & 0.034 & 0.175 & 0.357 \\
    \rowcolor{gray!15}
    SimPB*                               & ResNet101 & $1408\times512$ & 0.288 & 0.541 & 0.276 & \textbf{0.065} & \textbf{0.219} & 0.388 \\
    \rowcolor{gray!15}
    SimPB++*                       & ResNet101 & $1408\times512$ & \textbf{0.295}& \textbf{0.550}& \textbf{0.286}& 0.048& 0.210& \textbf{0.402}\\
    \bottomrule
\end{tabular}
}

\label{tab:nus_val_2D}
\end{table*} 
\begin{table*}[t]
\centering
\caption{Comparisons on the Argoverse 2 validation dataset with a range of 150 meters.}

\resizebox{0.8\textwidth}{!}
{
\begin{tabular}{@{}l|c|c|cc|ccccc@{}}
    \toprule
    Method  & \ \, Backbone\ \,  & \ \, Resolution\ \,& \ mAP$\uparrow$\,  & \,CDS$\uparrow$\ & \,mATE\,$\downarrow$ & \,mASE\,$\downarrow$ & \,mAOE\,$\downarrow$  \\
    \midrule
    BEVStereo\cite{BEVStereo}    & V2-99 &$960\times 640$& 0.146& 0.104& 0.847& 0.397& 0.901  \\
    SOLOFusion\cite{SOLOFusion}  & V2-99 &$960\times 640$& 0.149& 0.106& 0.934& 0.425& 0.779  \\
    PETR\cite{PETR}              & V2-99 &$960\times 640$& 0.176& 0.122& 0.911& 0.339& 0.819  \\
    Sparse4Dv2\cite{Sparse4Dv2}  & V2-99 &$960\times 640$& 0.189& 0.134& 0.832& 0.343& 0.723  \\
    StreamPETR\cite{StreamPETR}  & V2-99 &$960\times 640$& 0.203& 0.146& 0.843& 0.321& 0.650  \\
    RayDN\cite{RayDN}            & V2-99 &$960\times 640$& 0.223& 0.161& 0.825& 0.325& 0.629  \\
    Far3D\cite{Far3D}            & V2-99 &$960\times 640$& 0.244& 0.181& 0.796& \textbf{0.304}& 0.538  \\
    \rowcolor{gray!15}
    SimPB\cite{SimPB}            & V2-99 &$960\times 640$& 0.246& 0.186& 0.772& 0.322& 0.525   \\
    \rowcolor{gray!15}
    SimPB++                      & V2-99 &$960\times 640$& \textbf{0.286}& \textbf{0.217}& \textbf{0.732}& 0.310& \textbf{0.499}   \\
    \bottomrule
\end{tabular}
}
\label{tab:av2_val}
\end{table*}

\subsection{Main Results}
\textbf{nuScenes dataset.}
We first evaluate the 3D and 2D performance of SimPB++ on the nuScenes validation set. 
As shown in the Table \ref{tab:nus_val}, SimPB++ achieves the best results in 3D object detection. 
In particular, compared with recent state-of-the-art methods, SimPB brings a clear performance advantage, while SimPB++ further improves over SimPB by 1.8\% mAP and 1.3\% NDS with the configuration of an input resolution of 704$\times$256 and a ResNet50 backbone. 
Under a larger-scale configuration with a resolution of 1408$\times$512 and ResNet101 backbone, it still achieves gains of 1.0\% mAP and 1.0\% NDS.
It is worth noting that there is a significant improvement in mAOE, which can be attributed to the newly introduced Propagating Denoising module with \textit{alpha}-angle loss that enables more precise estimation of object orientations.

We also evaluate the 2D detection performance of SimPB++ on the nuScenes validation set, as shown in the Table \ref{tab:nus_val_2D}. Compared with the previous version, SimPB++ achieves improvements of 1.0\% and 0.7\% AP under the two configurations, respectively.
These gains can be mainly attributed to the proposed Propagating Denoising and Auxiliary Branch, which facilitate coarse obstacle detection and enhance feature representation. Additionally, the Crop-and-Scale strategy improves the recall of distant objects.

\noindent
\textbf{Argoverse2 dataset.}
Following \cite{Far3D, RayDN}, we further evaluate the performance of the SimPB++ series on the Argoverse2 validation set. As shown in Table~\ref{tab:av2_val}, SimPB++ achieves competitive results compared to recent state-of-the-art methods, demonstrating the potential of the unified architecture for detecting distant objects in challenging long-range scenarios. Benefiting from the proposed Crop-and-Scale module, SimPB++ attains significant gains of 4\% mAP and 3.1\% CDS, highlighting its enhanced capability in long-range object detection tasks. A detailed analysis of distant target detection is presented in Section~\ref{sec:ablation}.

\subsection{Ablation Study}
\label{sec:ablation}
In the ablation experiments, following \cite{SparseBEV, RayDN, GeoBEV}, we train 24 epochs on the validation split of nuScenes using a ResNet50 backbone with a resolution $704 \times 256$.
 To explore long-range detection, we also conducted ablation experiments on the Argoverse2 validation dataset, with a 3-epoch training phase using a V2-99 backbone and a resolution of $960 \times 640$ resolution. Other training parameters are kept consistent unless otherwise specified.

\begin{table*}[t]
\centering
\caption{The ablation studies of different combination of multi-view 2D layer and 3D layer in hybrid decoder layer.}
\resizebox{0.95\textwidth}{!}
{
\begin{tabular}{@{}c|cc|c|cc|ccccc@{}}
    \toprule
    \ Index \ & \ 2D layers \ & \ 3D layers \ & \ Hybrid layers \ &\ mAP$\uparrow$\,  & \,NDS$\uparrow$\ & \,mATE\,$\downarrow$ & \,mASE\,$\downarrow$ & \,mAOE\,$\downarrow$ & \,mAVE\,$\downarrow$ & \,mAAE\,$\downarrow$ \\
    \midrule
    A      & 0 & 1& 6& 0.397 & 0.504 & 0.607 & 0.270 & 0.594 & \textbf{0.270} & 0.196 \\
    B      & 1 & 0& 6& 0.397 & 0.503 & 0.635 & 0.279 & 0.540 & 0.297 & 0.204 \\
    C      & 2 & 1& 2& 0.417 & 0.508 & 0.605 & 0.274 & 0.543 & 0.363 & 0.212 \\
    D      & 1 & 2& 2& 0.419 & 0.517 & 0.599 & \textbf{0.269} & 0.555 & 0.300 & 0.206 \\
    E      & 3 & 3& 1& 0.419 & 0.523 & 0.595 & 0.270 & 0.526 & 0.277 & \textbf{0.192} \\
    \rowcolor{gray!15}
    F      & 1 & 1& 3& \textbf{0.421} & \textbf{0.527} & \textbf{0.590} & 0.274 & \textbf{0.492} & 0.287 & 0.195 \\
    \bottomrule
\end{tabular}
}
\label{tab:ablation_structure}
\end{table*}

\noindent
\textbf{Choice of Hybrid Decoder.}
The hybrid decoder in our proposed approach is designed with a total number of layers given by $L_{\text{total}} = (L_{\text{2d}} + L_{\text{3d}}) \times L_{\text{hybrid}}$. To ensure a fair comparison among different numbers of layers within the hybrid decoder, we limit the total number of layers to 6 by setting $L_{\text{total}} = 6$. The results of our ablation study are summarized in Table \ref{tab:ablation_structure}.

In experiment \textbf{A}, when $L_{\text{2d}} = 0$, the hybrid decoder becomes a standard 3D decoder module. In experiment \textbf{B}, when $L_{\text{3d}} = 0$, we have only 2D decoder layers with 3D object deep supervision after the Adaptive Query Aggregation block. It shows that, even in these degraded settings, our proposed decoder approach still achieves decent 3D object detection results. 
Interestingly, the model in experiment \textbf{B}, which exclusively utilizes multi-view 2D decoder layers with 3D auxiliary supervision, demonstrates comparable performance to experiment \textbf{A}. This observation emphasizes the effectiveness of our Dynamic Query Allocation and Adaptive Query Aggregation blocks in the proposed multi-view 2D decoder.
By incorporating both 2D decoder and 3D decoder layers (experiments \textbf{C}, \textbf{D}, \textbf{E}, and \textbf{F}), we achieve improved performance compared to the degraded settings.
In experiment \textbf{E}, we also simulate the two-stage scheme of a cascade 2D detector and 3D detector by setting $L_{\text{2d}} = 3$ and $L_{\text{3d}} = 3$. In the two-stage method, the 2D detector is usually applied to provide initialization of 3D queries. This approach yields superior results compared to model \textbf{A}, which lacks additional 3D query initialization. 
Therefore, the utilization of the multi-view 2D decoder aids in providing more accurate 3D query initialization for the subsequent 3D decoder in the hybrid decoder.
Lastly, model \textbf{F}, which is used in SimPB, employs a cyclic interaction between multi-view 2D and 3D layers and provides the best performance in terms of both mAP and NDS.
In summary, our hybrid decoder successfully exchanges information learned during both 2D and 3D object detection tasks, leading to improved performance of multi-view 3D detection.

\begin{table}
\centering
\caption{Ablation studies on query allocation mechanism.}
\resizebox{0.82\textwidth}{!}{
\begin{tabular}{l|cc}
    \toprule
    Allocation Strategy & \ mAP$\uparrow$\,  & \,NDS$\uparrow$\  \\
    \midrule
    Uniform                               & 0.365    & 0.474        \\
    Object Center                         & 0.410    & 0.515        \\
    Object Center + Front-Rear Point      & 0.414    & 0.520        \\
    \rowcolor{gray!15}
    Object Center + Anchor Corner \qquad  & \textbf{0.421}    & \textbf{0.527}        \\
    \bottomrule
\end{tabular}
}
\label{tab:ablation_allocation}
\end{table}

\noindent
\textbf{Effect of Dynamic Query Allocation Strategy.} 
In Table \ref{tab:ablation_allocation}, we compare different ways for query allocation.
A simple way is to uniformly distribute 3D queries to different cameras without camera information and generate 2D queries. It employs a fixed number of queries for both 2D and 3D objects, without considering the varying number of targets in different cameras. As a result, it overlooks the potential benefits of utilizing 2D objects in the image view. 
We propose three different ways to dynamically allocate 3D anchors to their corresponding cameras using camera parameters. This includes projecting a single object center, an object center with front and rear face centers, and an object center with eight corner points.
These dynamic query allocation methods yield superior results compared to the uniform method. 
In multi-view object detection, objects can appear across different cameras. Therefore, projecting only the single object center to one camera may lead to inaccurate localization of truncated objects in other cameras. To address this limitation, we propose projecting additional points such as the front and rear face centers of an object, as well as the eight corner points of a 3D box. These approaches yield improved results compared to using only a single object center. In summary, by utilizing the object center and eight corners of a 3D box, our method achieves the best results.

\begin{table}
\centering
\caption{Ablation studies on query aggregation mechanism.}
\resizebox{0.8\textwidth}{!}{
\begin{tabular}{l|cc}
    \toprule
    Aggregation Strategy \ & \ mAP$\uparrow$\,  & \,NDS$\uparrow$\  \\
    \midrule
    Cross-Attention                \  & 0.365     & 0.474    \\
    Cross-Attention + Supervision  \  & 0.399     & 0.495    \\
    Query Fusion                   \  & 0.389     & 0.489    \\
    \rowcolor{gray!15}
    Query Fusion + Supervision     \  & \textbf{0.421}     & \textbf{0.527}    \\
    \bottomrule
\end{tabular}
}
\label{tab:ablation_aggregation}
\end{table}

\noindent
\textbf{Effect of Adaptive Query Aggregation Strategy.} 
In Table \ref{tab:ablation_aggregation}, we explore different methods to fuse 2D queries $\tilde{Q}_{\text{2d}}$ and 3D queries $Q_{\text{3d}}$. We compare our proposed query fusion approach with a simple cross-attention fusion method. Additionally, we investigate the impact of applying deep supervision to the fused 3D queries $Q_{\text{3d}}^{\text{agg}}$ by using 3D ground truth.
The results show that our query fusion approach outperforms the simple cross-attention fusion method. This finding validates the effectiveness of our Adaptive Query Aggregation block in successfully fusing 2D queries to construct 3D queries for the subsequent layer.
Furthermore, incorporating 3D deep supervision for the aggregated 3D queries in the multi-view 2D decoder enhances the detection performance for both methods.

\begin{table}
\centering
\caption{Ablation studies on the effectiveness of the enhancement strategies including Crop-and-Scale (CAS), Propagating Denosing (PDN), and Auxiliary Branch (AUX).}
\resizebox{0.9\textwidth}{!}
{
\begin{tabular}{ccc|cc|cc}
    \toprule
    \multirow{2}{*}{CAS} & \multirow{2}{*}{PDN} & \multirow{2}{*}{AUX} & \multicolumn{2}{c|}{\textbf{nuScenes}} & \multicolumn{2}{c}{\textbf{Argoverse2}} \\
    &     &     & \ mAP$\uparrow$\,   & \,NDS$\uparrow$\  & \ mAP$\uparrow$\,  & CDS $\uparrow$ \\
    \midrule
               &            &            & 0.421 & 0.527 &  0.202 & 0.149 \\
    \checkmark &            &            & 0.432 & 0.533 &  0.217 & 0.162  \\
    \checkmark & \checkmark &            & 0.446 & 0.546 &  0.224 & 0.165 \\
   \rowcolor{gray!15}
   \checkmark & \checkmark & \checkmark & \textbf{0.460} & \textbf{0.555} &  \textbf{0.233} & \textbf{0.172} \\

    \bottomrule
\end{tabular}
}
\label{tab:ablation_enhancement}
\end{table}

\noindent
\textbf{Ablation on Enhancement Strategies.} 
To verify the effectiveness of enhancement strategies, including Scale-and-Crop (SAC), Propagating Denosing (PDN), and Auxiliary Branch (AUX), we conduct ablation experiments on nuScenes and Argoverse2.
As shown in Table~\ref{tab:ablation_enhancement}, we sequentially add modules to extend the baseline model SimPB into SimPB++ from top to bottom.
It can be observed that adding each strategy provides a positive return compared to the previous variation.
Employing the Scale-and-Crop strategy yields improvements of 0.15\% in mAP and 0.13\% in CDS on Argoverse2, as it is compatible with the evaluation of long-range objects. Meanwhile, utilizing the propagating denoising strategy enhances mAP by 0.14\% and NDS by 0.13\% on the nuScenes dataset, facilitating faster convergence. 
Additionally, the auxiliary branch strengthens foreground features with depth estimation before the decoder, achieving a new record.
There is a notable improvement in integrating three strategies compared with the baseline model.
Only 24 epoch training with about 5 hours can achieve comparable score with previous state-of-art works on nuScenes dataset.

\begin{table}
\centering
\caption{Ablation studies on the additional views of Scale-and-Crop Strategy.}
\resizebox{0.995\textwidth}{!}
{
\begin{tabular}{c|ccc|ccc}
    \toprule
    \multirow{2}{*}{Additional Views} & \multicolumn{3}{c|}{\textbf{nuScenes}} & \multicolumn{3}{c}{\textbf{Argoverse2}} \\
    & \ mAP$\uparrow$\,   & \,NDS$\uparrow$\  & \,FPS$\uparrow$\  & \ mAP$\uparrow$\,  & CDS $\uparrow$ & \,FPS$\uparrow$\ \\
    \midrule
   -                   & 0.448 & 0.545 & 17.0 & 0.216 & 0.160 & 6.2 \\
   \rowcolor{gray!15}
   Front\,\&\,Fear     & 0.460 & 0.555 & 15.6 & 0.233 & 0.172 & 4.9 \\
   All Surroundings    & 0.473 & 0.567 & 14.4 & 0.249 & 0.184 & 4.0 \\

    \bottomrule
\end{tabular}
}
\label{tab:ablation_scale_crop_view}
\end{table}

\begin{table}
\centering
\caption{Ablation studies on the scale rate of Scale-and-Crop Strategy.}
\resizebox{0.74\textwidth}{!}
{
\begin{tabular}{c|cc|cc}
    \toprule
    \multirow{2}{*}{Scale Rate} & \multicolumn{2}{c|}{\textbf{nuScenes}} & \multicolumn{2}{c}{\textbf{Argoverse2}} \\
    & \ mAP$\uparrow$\,   & \,NDS$\uparrow$\  & \ mAP$\uparrow$\,  & CDS $\uparrow$ \\
    \midrule
   1.5        & 0.455 & 0.553 & 0.232 & 0.172 \\
   \rowcolor{gray!15}
   2.0        & 0.460 & 0.555 & 0.233 & 0.172 \\
   2.5        & 0.458 & 0.555 & 0.231 & 0.171 \\

    \bottomrule
\end{tabular}
}
\label{tab:ablation_scale_crop_rate}
\end{table}

\noindent
\textbf{Effect of Crop-and-Scale strategy.}
We conduct two ablation studies to evaluate the proposed Crop-and-Scale strategy: one examining the impact of incorporating additional views, and another analyzing the effect of different scale rates.
Table~\ref{tab:ablation_scale_crop_view} compares the strategy with different combinations of additional perspective views.
Compared to the baseline, enabling the Crop-and-Scale strategy with additional front and rear views improves performance by supplying fine-grained information. Extending the approach to include all surrounding views yields further gains. However, we adopt the front-rear configuration as the default for two reasons: first, front and rear views contain more task-relevant information than side views, which predominantly capture street scenes and buildings; second, incorporating all surrounding views increases training time and reduces inference efficiency. A detailed analysis of training with all surrounding views is provided in the Appendix.

Table~\ref{tab:ablation_scale_crop_rate} presents the results of the Crop-and-Scale strategy with different scale rates (1.5, 2.0, and 2.5). The comparable performance across these settings demonstrates the model's strong generalization ability. We select a scale rate of 2.0 as the default, as it achieves the best overall performance.

\begin{table}[!t]
\centering
\caption{Ablation studies on the Propagating Denoising.}
\resizebox{0.95\textwidth}{!}
{
\begin{tabular}{cc|cc|cc}
    \toprule
    \multirow{2}{*}{\textbf{3D-denosing}} & \multirow{2}{*}{\textbf{2D-denosing}} & \multicolumn{2}{c|}{\textbf{nuScenes}} & \multicolumn{2}{c}{\textbf{Argoverse2}} \\
    &  & \ mAP$\uparrow$\,   & \,NDS$\uparrow$\  & \ mAP$\uparrow$\,  & CDS $\uparrow$ \\
    \midrule
     &                      & 0.434 & 0.537 & 0.224 & 0.163 \\
    \checkmark &            & 0.452 & 0.548 & 0.230 & 0.168 \\
   \rowcolor{gray!15}
    \checkmark & \checkmark & 0.460 & 0.555 & 0.233 & 0.172 \\
   \bottomrule
\end{tabular}
}
\label{tab:ablation_propagating_denoising}
\end{table}

\noindent
\textbf{Effect of Propagating Denoising.}
The Propagating Denoising module integrates joint denoising of 3D and 2D tasks, where 2D denoising is an extension built upon the 3D denoising foundation. To disentangle the contribution of each component, we conduct ablation experiments by enabling the 3D and 2D denoising modules separately within the Propagating Denoising framework.

As shown in Table~\ref{tab:ablation_propagating_denoising}, activating the 3D denoising module alone yields a substantial improvement in model performance, confirming the effectiveness of noise reduction in enhancing feature quality. When 2D denoising is further incorporated to activate the full Propagating Denoising mechanism, the model achieves an additional performance gain. This incremental improvement highlights the synergistic effect between the 3D and 2D denoising components: the 3D module establishes a robust foundation for spatial structure refinement, while the 2D extension complements it by capturing fine-grained texture details, collectively strengthening the model's ability to process noisy inputs.

\begin{table}
\centering
\caption{Ablation studies on the Auxiliary Branch.}
\resizebox{1\textwidth}{!}
{
\begin{tabular}{ccc|cc|cc}
    \toprule
    \multirow{2}{*}{\textbf{Map DepthNet}}  &
    \multirow{2}{*}{\textbf{RoI Detector}}  &
    \multirow{2}{*}{\textbf{Ins. DepthNet}} &
    \multicolumn{2}{c|}{\textbf{nuScenes}}  &
    \multicolumn{2}{c}{\textbf{Argoverse2}} \\
    &  & & \ mAP$\uparrow$\,   & \,NDS$\uparrow$\  & \ mAP$\uparrow$\,  & CDS $\uparrow$ \\
    \midrule
   &  &                                 & 0.446 & 0.546 & 0.209 & 0.155 \\
   \checkmark &  &                      & 0.451 & 0.548 & 0.217 & 0.159 \\
   \checkmark &  \checkmark&            & 0.453 & 0.550 & 0.227 & 0.168 \\
   \rowcolor{gray!15}
   \checkmark & \checkmark & \checkmark & 0.460 & 0.555 & 0.233 & 0.172  \\

    \bottomrule
\end{tabular}
}
\label{tab:ablation_auxiliary_branch}
\end{table}

\begin{table}
\centering
\caption{The ablation studies of camera observation loss.}
\resizebox{0.59\textwidth}{!}{
\begin{tabular}{@{}l|ccc@{}}
    \toprule
    \ \ $\lambda_{alpha}$   & \ mAP$\uparrow$\,  & \,NDS$\uparrow$ \, & \ mAOE$\downarrow$ \\
    \midrule
    \ \ -            & 0.421          & 0.514 & 0.611  \\
    \ \ 0.25         & \textbf{0.425} & 0.519 & 0.582  \\
    \rowcolor{gray!15}
    \ \ 0.50         & 0.421 & \textbf{0.527} & 0.492  \\
    \ \ 1.00         & 0.418 & 0.523 & \textbf{0.465}  \\
    
    \bottomrule
\end{tabular}
}
\label{tab:ablation_loss}
\end{table}

\noindent
\textbf{Impact of observation angle loss.} 
We also investigate the influence of the observation angle loss $\mathcal{L}_{\text{alpha}}$. By incorporating $\mathcal{L}_{\text{alpha}}$ from the image view into the training loss, we observe a slight improvement in mAP and NDS, along with a significant decrease in the mean average orientation error (mAOE). Increasing the weight of $\lambda_{\text{alpha}}$ leads to a decrease in mAOE. $\lambda_{\text{alpha}}=0.5$ is used in our SimPB++.

\begin{figure*}[t]
  \centering
  \includegraphics[width=0.95\linewidth]{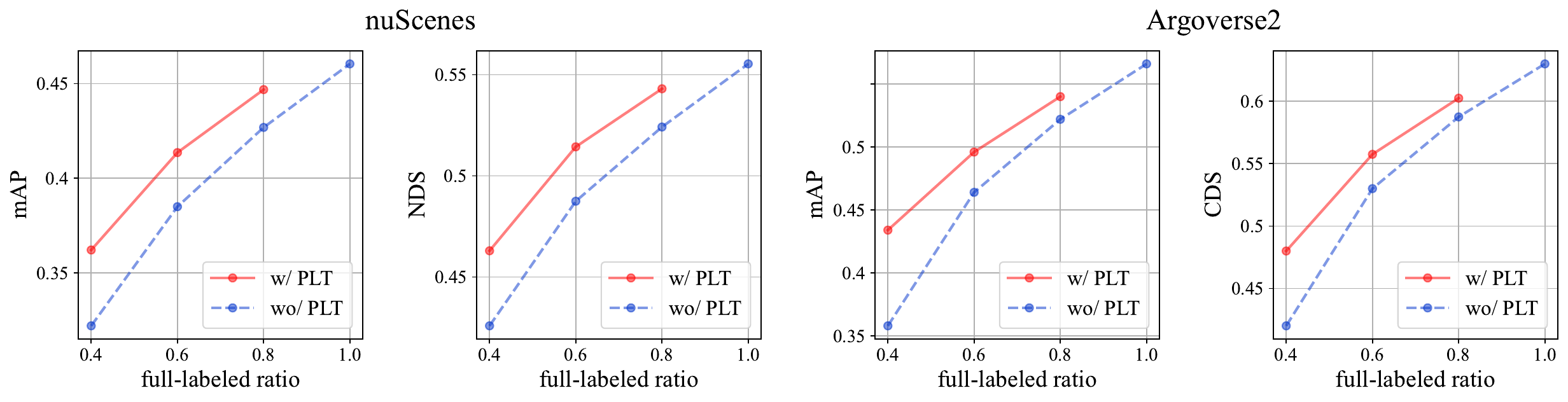}
  \caption{Performance analysis of Partial Labeling Training (PLT) with additional 2D-only labeled data for training.}
  \label{fig:partial_labeling_curve}
\end{figure*}
\noindent

\subsection{Analysis and Discussion}

\begin{figure}[t]
  \centering
  \includegraphics[width=0.95\linewidth]{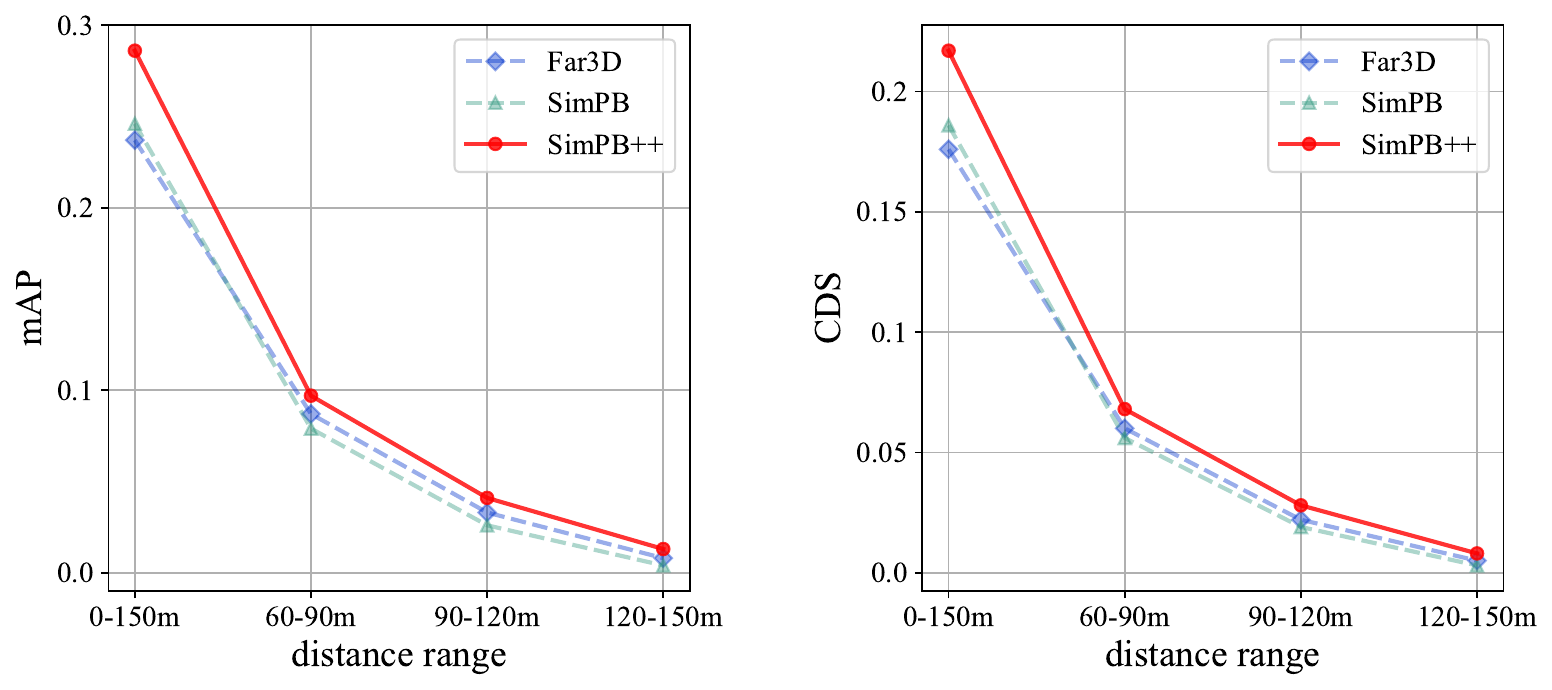}
  \caption{Comparison of long-range detection on Argoverse2.}
  \label{fig:long_range_curve}
\end{figure}

\noindent
\textbf{Analysis of Long-range Detection.}

To evaluate the long-range detection capability of our approach, we conduct distance-segmented experiments on the Argoverse2 dataset, comparing SimPB++ against both its preliminary version SimPB~\cite{SimPB} and Far3D~\cite{Far3D}, a method specifically designed for long-range object detection.

As shown in Figure~\ref{fig:long_range_curve}, we group objects into four distance intervals: a global range (0–150 m) and three long-range intervals (60–90m, 90–120m, and 120–150m). Our previous version, SimPB, achieves slightly better performance than Far3D over the global range but falls short in the long-range intervals, where it is marginally outperformed by Far3D.
In contrast, SimPB++ demonstrates comprehensive superiority: it not only preserves the strengths of SimPB in short-range detection but also excels in long-range scenarios, achieving enhanced overall performance across all distance ranges.

\noindent
\textbf{Analysis of Partial-labeling Training.}

It is widely recognized that 3D annotation is significantly more difficult and expensive than 2D labeling. Given that the SimPB++ framework integrates both 2D and 3D capabilities, it can naturally accommodate training with either type of annotation. This raises an important question: \textit{Can we leverage low-cost 2D labels to improve 3D detection performance within SimPB++?}

To investigate this, we conduct experiments on the nuScenes dataset~\cite{Nuscene}. We compare two training strategies: (1) training with varying proportions of fully annotated data (containing both 2D and 3D labels), and (2) partial-labeling training, where we augment the fully annotated data with additional 2D-only labeled samples, with the proportion of such partial annotations fixed at 0.2.
As shown in Figure~\ref{fig:partial_labeling_curve}, with additional partial-labeling 2D data in the training consistently improves 3D detection performance. Notably, when compared to training with the same amount of fully annotated data, the partial-labeling method reaches the performance of models trained exclusively on full 3D annotations. These results demonstrate that the unified architecture of SimPB++ can effectively mitigate the scarcity of 3D annotation data by incorporating readily available 2D labels to enhance model performance.

\begin{table}
\centering
\caption{The analysis of the efficiency on a NVIDIA 4090 GPU. CAS indicates the Crop-and-Scale strategy.}
\resizebox{0.95\textwidth}{!}
{
\begin{tabular}{l|l|cc|c|c}
    \toprule
  Dataset &  Method   & Backbone & Resolution & CAS & FPS \\
    \midrule
    &SimPB      & ResNet50 & $704 \times 256$ &            &12.0 \\
    &SimPB++    & ResNet50 & $704 \times 256$ &            &16.0 \\
    \rowcolor{gray!15}
    \cellcolor{white}
    &SimPB++    & ResNet50 & $704 \times 256$ & \checkmark &15.6 \\
    \cmidrule{2-6} 
    &SimPB      & ResNet101 & $1408 \times 512$ &            &7.8 \\
    &SimPB++    & ResNet101 & $1408 \times 512$ &            &10.5 \\
    \rowcolor{gray!15}
    \cellcolor{white}
    \multirow{-6}{*}{nuScenes}
    &SimPB++    & ResNet101 & $1408 \times 512$ & \checkmark &9.2 \\
    \midrule
    &SimPB      & V2-99 & $960 \times 640$ &                &4.6 \\
    &SimPB++    & V2-99 & $960 \times 640$ &                &6.4 \\
    \rowcolor{gray!15}
    \cellcolor{white}
    \multirow{-3}{*}{Argoverse2}
    &SimPB++    & V2-99 & $960 \times 640$ & \checkmark     &4.9 \\
    \bottomrule
\end{tabular}
}
\label{tab:analysis_efficiency}
\end{table}

\noindent
\textbf{Analysis of Inference Efficiency.}
In our previous version, inference efficiency is constrained by the dynamic query allocation module, which relies on iterative loops across multiple camera views. To address this limitation, we optimize the process by restructuring the allocation into parallel operations, enabling concurrent execution across batch sizes and camera views. Additionally, we remove the encoder layer in low-resolution configurations to further speed up inference.
As reported in Table~\ref{tab:analysis_efficiency}, SimPB++ reduces inference time compared to SimPB, both with and without the Crop-and-Scale strategy. The Crop-and-Scale module introduces distant camera views as additional inputs to enhance long-range detection. This addition has a negligible impact on lightweight backbones and low-resolution settings, with FPS decreasing only slightly from 16.0 to 15.6. For heavier backbones, the reduction remains acceptable, at 1.3 and 1.5 FPS, respectively. Overall, SimPB++ maintains a favorable speed advantage over other state-of-the-art methods.

\begin{figure*}[t]
  \centering
  \includegraphics[width=0.99\linewidth]{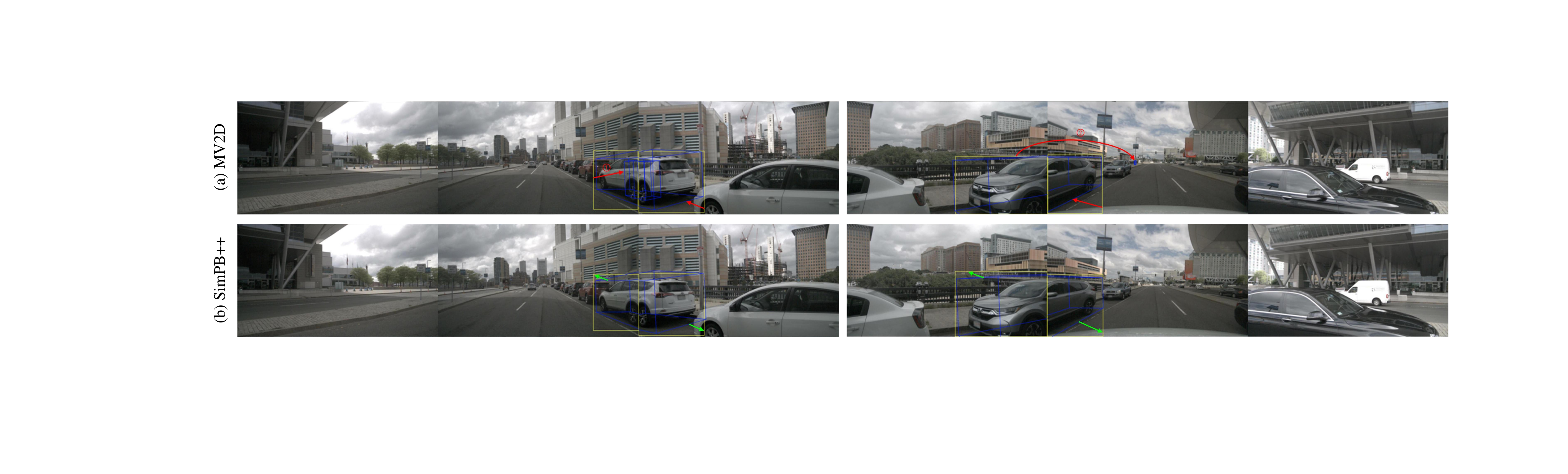}
  \caption{Comparison of association strategies. (a) MV2D (2D-to-3D association): \hlr{Red arrows} indicate duplicate or unrelated 3D results derived from 2D priors. (b) SimPB++ (3D-to-2D association): \hlg{Green arrows} show accurate 3D-2D correspondence for cross-camera targets, showcasing superior association accuracy. }
  \label{fig:association_comparision}
\end{figure*}
\noindent

\subsection{Qualitative Evaluation}
\label{sec:qualitative}

To conduct a qualitative analysis of the association establishment between 2D-to-3D and 3D-to-2D, we compare the detection results of MV2D and SimPB++ in the same keyframe. 
The detection results of several cross-camera targets are shown in Figure~\ref{fig:association_comparision}, where the yellow boxes represent the 2D detection results, and the blue boxes represent the 3D detection results.

In the case of a cross-camera target $O$, MV2D initially employs a 2D detector to generate multiple 2D bounding boxes (using two as an example). 
These 2D results are used to initialize 3D queries through a 2D-to-3D association method. However, multiple 3D queries are associated with the target $O$. Only one of these 3D queries accurately predicts the target, while the other may produce a duplicated nearby object ({\color{red}\textcircled{1}} in Figure~\ref{fig:association_comparision} (a)) or even an unrelated result ({\color{red}\textcircled{2}} in Figure~\ref{fig:association_comparision} (b))).
This discrepancy arises during the Hungarian matching step, where only the best candidate query is optimized as the positive sample, resulting in the suppression of the remaining 3D queries. Consequently, the 2D information from a specific view of the suppressed 3D query is discarded, despite it can provide relevant information about the same target.

To address this issue, SimPB++ adopts a novel approach to establish the association between 2D and 3D results using a 3D-to-2D method. For a cross-camera target, we distribute its 3D queries to different views for 2D detection tasks and subsequently aggregate the results to form a single 3D query.
To this end, for a cross-camera object, SimPB++ only produces one 3D detection result along with its corresponding 2D detection in each relevant camera. Also, our cyclic 3D-2D-3D interaction ensures there is a single coherent representation of the target across different views, eliminating redundancy outputs and enhancing the accuracy of the results (in Figure~\ref{fig:association_comparision} (b))).

\subsubsection{Qualitative Comparison with State-of-the-Arts}

\begin{figure}[!t]
\centering
\includegraphics[width=\linewidth]{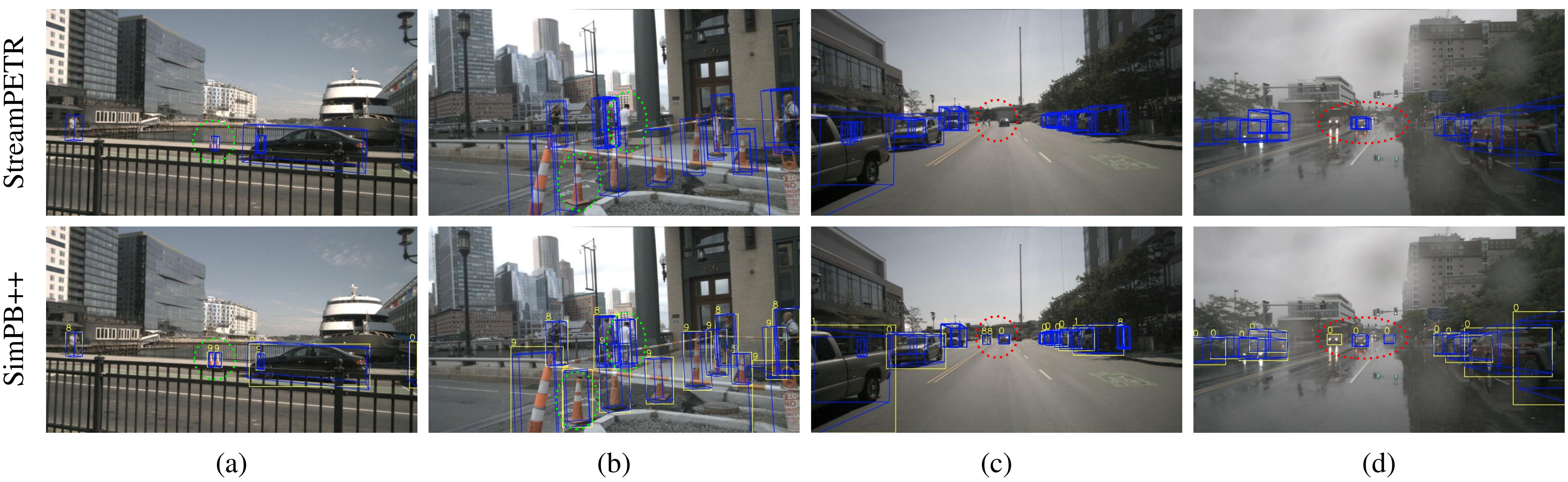}
\caption{Visualization results of StreamPETR and SimPB++.}
\label{fig:compare_with_streampetr}
\end{figure}

SimPB++ provides improved accuracy in detecting crowd objects such as traffic cones and pedestrians compared to StreamPETR~\cite{StreamPETR}. For instance, while StreamPETR incorrectly identifies two traffic cones as a single entity, SimPB++ accurately detects them as separate objects (green circle in Figure \ref{fig:compare_with_streampetr}~(a)).
In Figure \ref{fig:compare_with_streampetr}~(b), StreamPETR also provides an inaccurate estimation of the locations of pedestrians and traffic cones, showing that crowd objects tend to cluster around their neighboring objects. In contrast, SimPB++ provides more precise results and successfully distinguishes crowded and small objects. This improvement can be attributed to the novel cyclic 3D-2D-3D scheme of SimPB++, where the iterative and interactive process of 2D and 3D information enhances the refinement of queries, resulting in more accurate detection results.

SimPB++ also demonstrates its advantage in detecting distant targets and performs well even in challenging scenarios. 
For example, SimPB++ successfully detects pedestrians and cars at far distances, whereas StreamPETR fails to do so (red circle in Figure \ref{fig:compare_with_streampetr}~(c)).
Furthermore, despite encountering difficulties in predicting small and distant targets within complex environments, such as rain (as shown in \ref{fig:compare_with_streampetr}~(d)), SimPB++ can still provide reliable 2D detections. These 2D detections can be utilized in subsequent post-processing steps within a practical autonomous driving perception system.

\subsubsection{More Visualization Results}

\begin{figure*}[!htp]
\centering
\begin{minipage}{0.5\linewidth}
\centering
\includegraphics[width=0.98\linewidth]{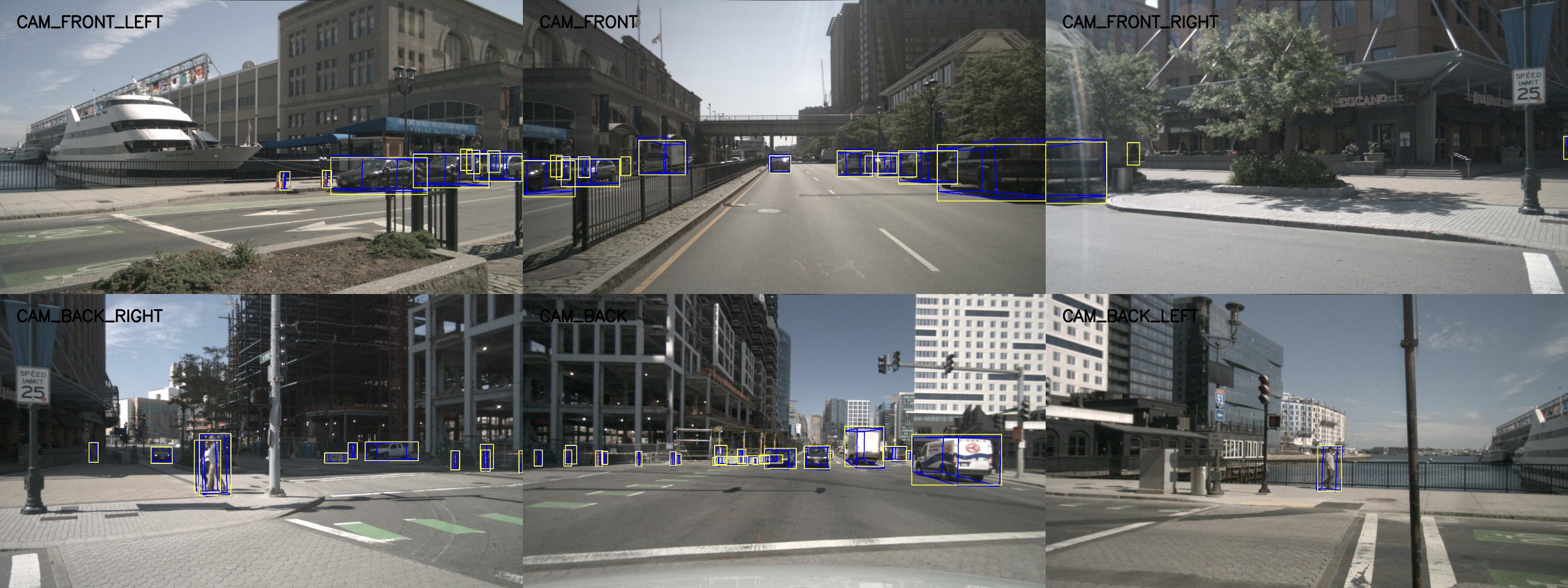}
\end{minipage}%
\begin{minipage}{0.5\linewidth}
\centering
\includegraphics[width=0.98\linewidth]{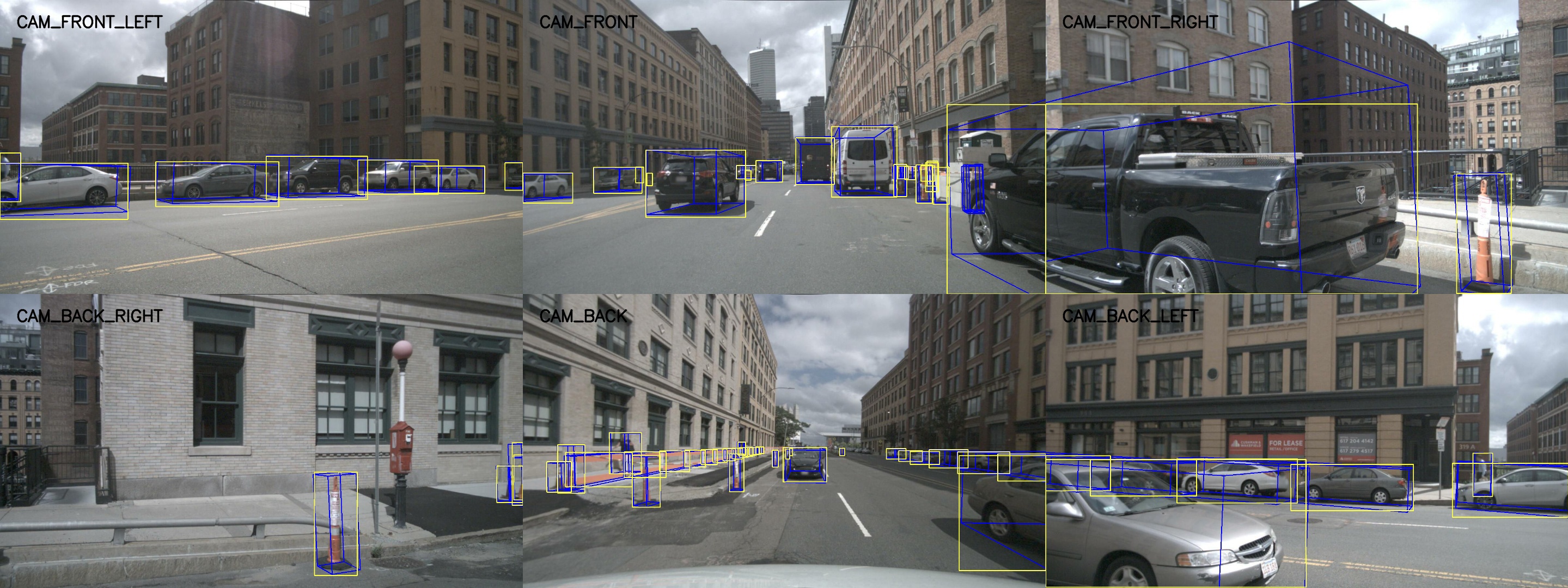}
\end{minipage}

\vspace{5pt}

\begin{minipage}{0.5\linewidth}
\centering
\includegraphics[width=0.98\linewidth]{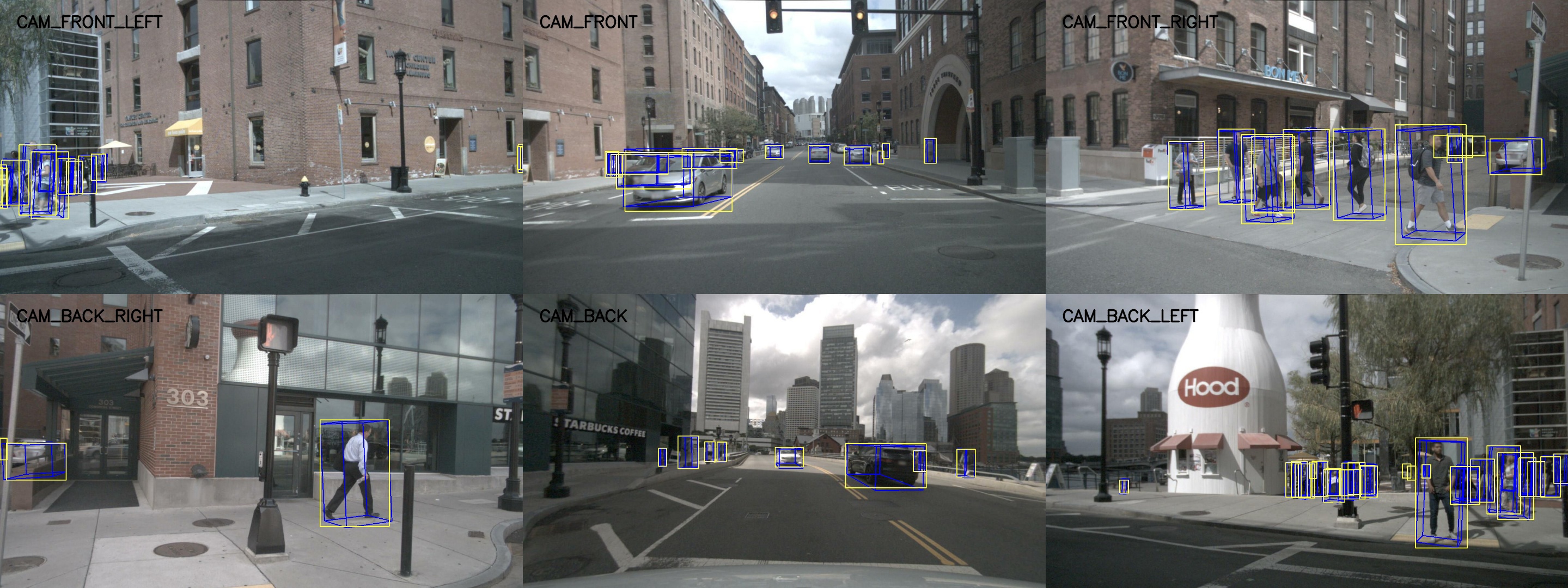}
\end{minipage}%
\begin{minipage}{0.5\linewidth}
\centering
\includegraphics[width=0.98\linewidth]{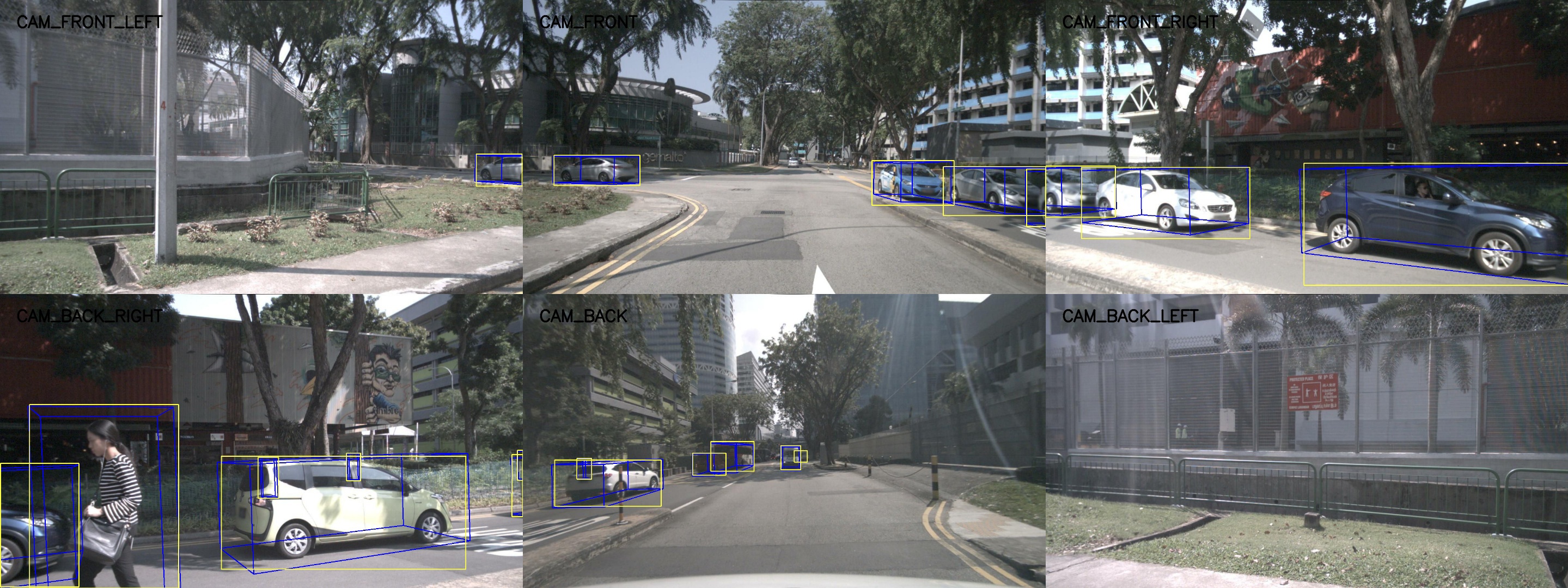}
\end{minipage}

\caption{Detection results on the nuScenes validation dataset.}
\label{fig:sup_viusal_nus}
\end{figure*}

\begin{figure*}[!htp]
\centering
\begin{minipage}{0.5\linewidth}
\centering
\includegraphics[width=0.98\linewidth]{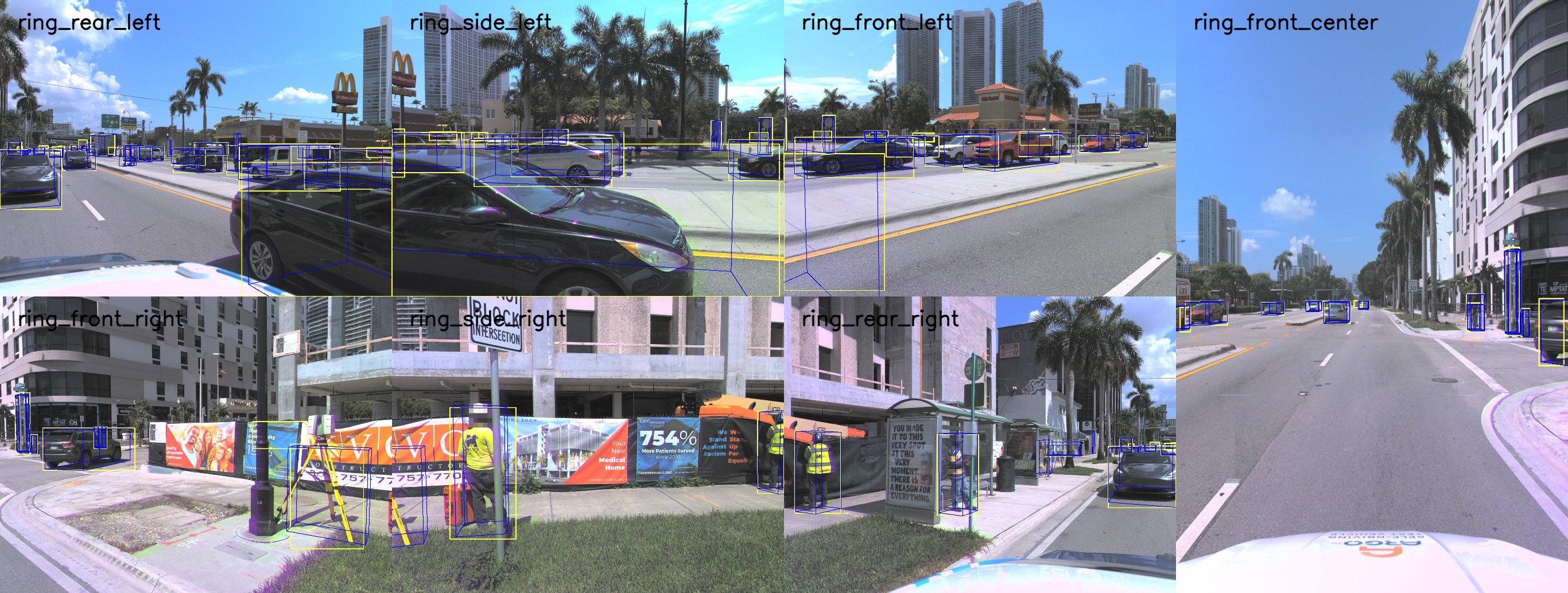}
\end{minipage}%
\begin{minipage}{0.5\linewidth}
\centering
\includegraphics[width=0.98\linewidth]{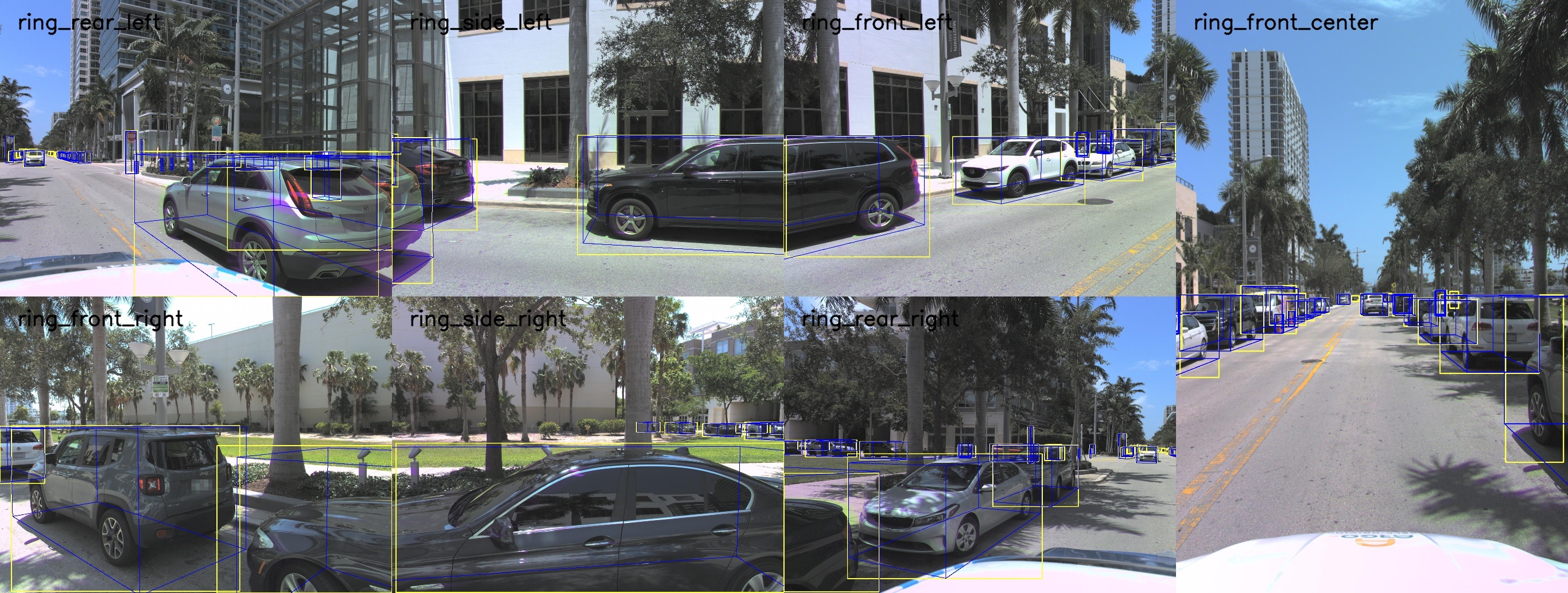}
\end{minipage}

\vspace{5pt}

\begin{minipage}{0.5\linewidth}
\centering
\includegraphics[width=0.98\linewidth]{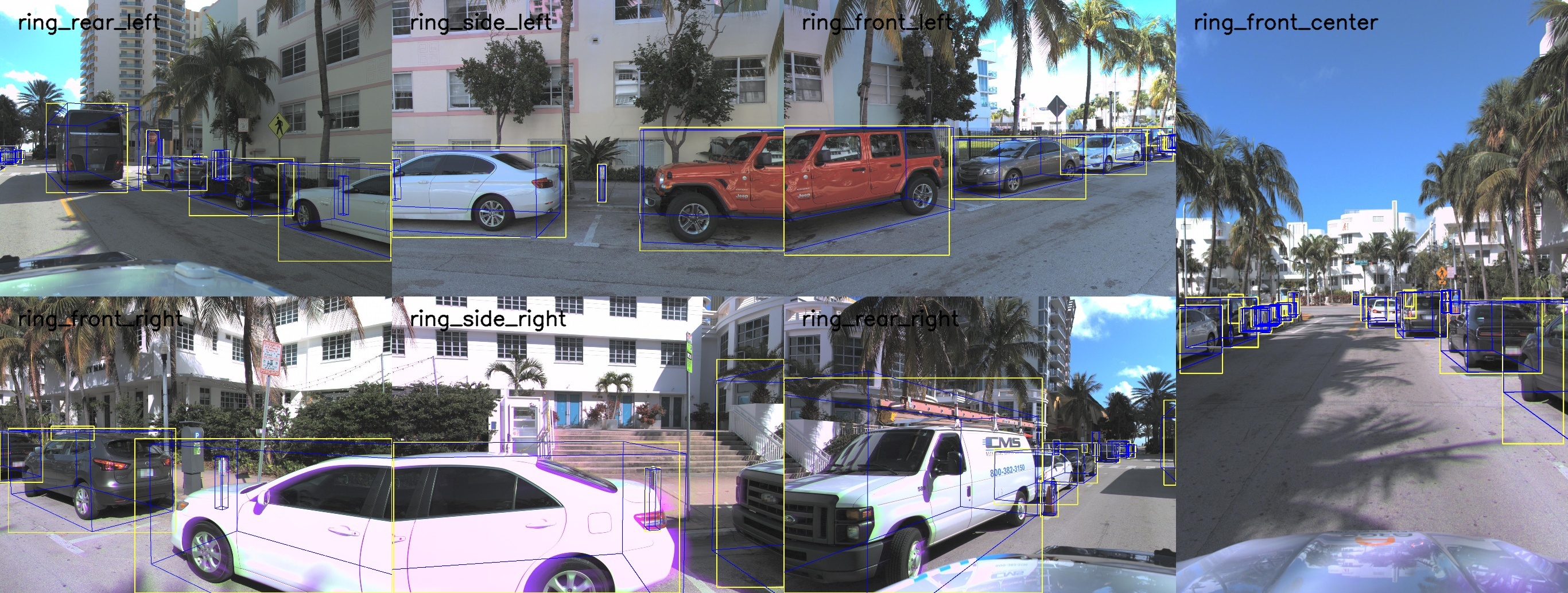}
\end{minipage}%
\begin{minipage}{0.5\linewidth}
\centering
\includegraphics[width=0.98\linewidth]{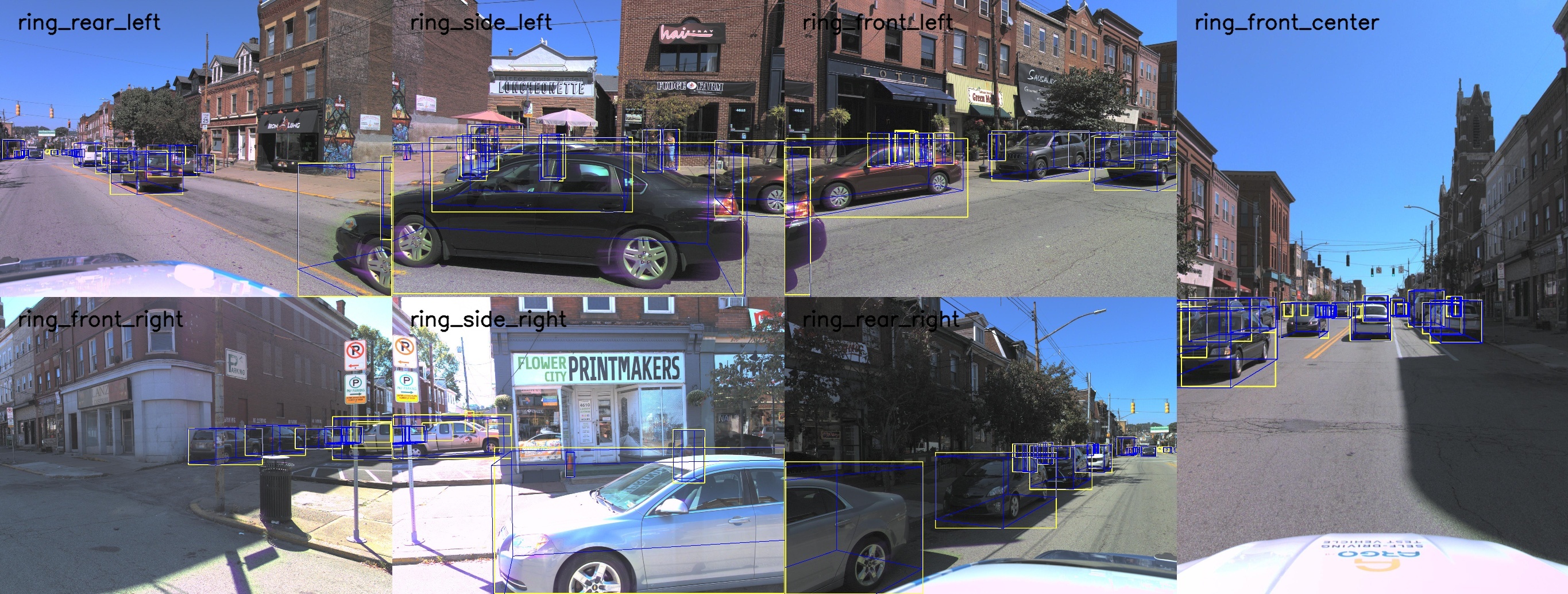}
\end{minipage}

\caption{Detection results on the Argoverse2 validation dataset.}
\label{fig:sup_viusal_av2}
\end{figure*}

The visualizations of the 2D and 3D detection results of SimPB++ are depicted in Figure \ref{fig:sup_viusal_nus} and Figure \ref{fig:sup_viusal_av2}, where the 2D predicted results are highlighted in \hly{yellow} and the 3D results are represented in \hlb{blue}.

\vspace{8pt}
\section{Conclusion and Limitation}

This paper presents a single-stage query-based framework for unified 2D and 3D multi-view detection. We introduce a Hybrid Decoder that interleaves 2D and 3D  layers for image-space and BEV-space, facilitated by dynamic allocation and adaptive aggregation modules for seamless cyclic 2D-3D interaction. 
To enhance perception, query-group attention refines intra-camera interactions, while a crop-and-scale strategy optimizes long-range detection via dynamic sampling. 
Performance is further improved by a propagating denoising strategy for unified query refinement and an auxiliary branch (ROI and depth-guided) to fortify foreground features. 
Finally, we demonstrate that partial label training significantly alleviates the 3D annotation burden, offering a cost-effective solution for large-scale autonomous driving.
We thoroughly evaluate SimPB++ on the nuScenes dataset with extensive experiments for both 2D and 3D tasks, as well as on the Argoverse2 dataset for distant 3D detection.

\bibliographystyle{IEEEtran}
\bibliography{main}

\clearpage

\appendices
\setcounter{table}{0}\refstepcounter{section}

\setcounter{page}{1}

\twocolumn[
  \begin{@twocolumnfalse}
    \begin{center}
      {\huge \textbf{Supplementary Material}}
    \end{center}
    \vspace{1em}
  \end{@twocolumnfalse}
]

\renewcommand{\theequation}{S.\arabic{equation}}
\setcounter{equation}{0}
\renewcommand{\thefigure}{S.\arabic{figure}}
\setcounter{figure}{0}
\renewcommand{\thetable}{S.\arabic{table}}


\begin{figure*}[htpb]
\centering
\includegraphics[width=0.95\textwidth]{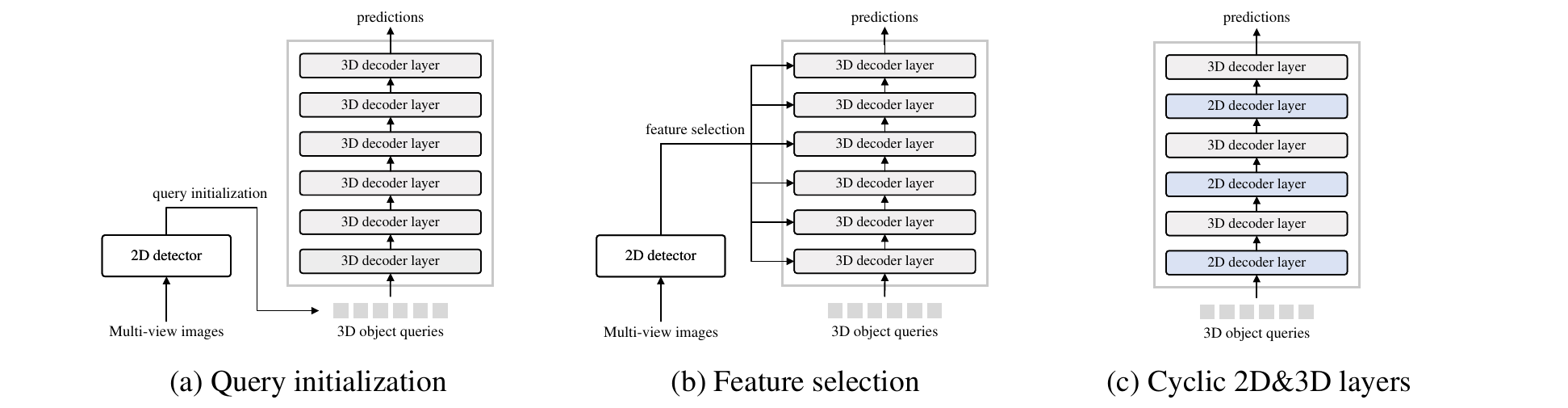}
\caption{Comparison of SimPB++ with previous approaches utilizing 2D results as priors. We roughly categorize these previous methods into two categories, (a) query initialization and (b) feature selection. Instead, SimPB++ introduces a unified paradigm using novel cyclic 2D\&3D layers as in (c).}
\label{fig:sup_difference_w_previous}
\end{figure*}

This supplementary material contains additional details of the main manuscript and provides more experiment analysis. 
In Section \ref{sec:difference}, we present the difference between SimPB++ and other previous works that utilize 2D results as priors. 
Next, we elaborate on the complete architecture and give more implementation details in Section \ref{sec:implementation}. 
Then, we provide an experiment analysis on the association accuracy in Section \ref{sec:experiment}. 
Finally, more visualization results are illustrated in Section \ref{sec:visualization}.

\subsection{Utilizing 2D Results as Priors}
\label{sec:difference}

We highlight the distinctions between SimPB++ and previous approaches that use 2D results as priors from two perspectives: architecture and association. A comprehensive comparison is provided in Figure \ref{fig:sup_difference_w_previous}.

\textbf{Architecture.} 
We categorize the previous methods into two groups: query initialization and feature selection, as summarized below.
\begin{itemize}
    \item Query Initialization: In this category, 3D queries are typically initialized from 2D boxes that are detected by a 2D detector (as shown in Figure \ref{fig:sup_difference_w_previous} (a)).
    \item Feature Selection: The methods focus on foreground tokens through 2D supervision and then select them for interaction with 3D queries (as shown in Figure \ref{fig:sup_difference_w_previous} (b)).
\end{itemize}
All these methods employ a 2D detector (or utilize a 2D head) to predict 2D results as a preliminary step before applying a 3D detector.
In contrast, SimPB++ takes a distinct approach. It performs simultaneous multi-view 2D and 3D detection within a single model using cyclic 2D \& 3D decoder layers (as illustrated in Figure \ref{fig:sup_difference_w_previous} (c)). SimPB++ is a one-stage method that does not rely on an off-the-shelf 2D detector.

\textbf{Association.} 
For association, we refer to the connection between 2D and 3D results for the same target. A summary of the association of previous methods is listed as follows.
\begin{itemize}
    \item Query Initialization: This method employs a heuristic default association, where a 3D query is linked to a 2D box for its initialization. This association is referred to as a 2D-to-3D association (as shown in Figure \ref{fig:sup_difference_w_previous} (a)).
    \item Feature Selection: In this approach, the association between 3D queries and selected 2D image tokens, supervised by a 2D detector, is learned through the transformer (as shown in Figure \ref{fig:sup_difference_w_previous} (b)). However, it does not explicitly establish a direct association between 2D and 3D results.
\end{itemize}
In contrast, our method determines the association by projecting 3D anchors and matching them with the corresponding 2D results (as illustrated in Figure \ref{fig:sup_difference_w_previous} (c)) . 
In this way, our approach establishes a 3D-to-2D association between 2D and 3D results. 
The 3D-to-2D association has the advantage of aggregating 2D information more efficiently and avoiding the generation of redundant 3D results. We give a detailed analysis in Section \ref{sec:effectiveness_association}.

\subsection{More Implementation Details}
\label{sec:implementation}
\subsubsection{Architecture Details}

\begin{figure*}[htbp]
\centering
\includegraphics[width=0.95\textwidth]{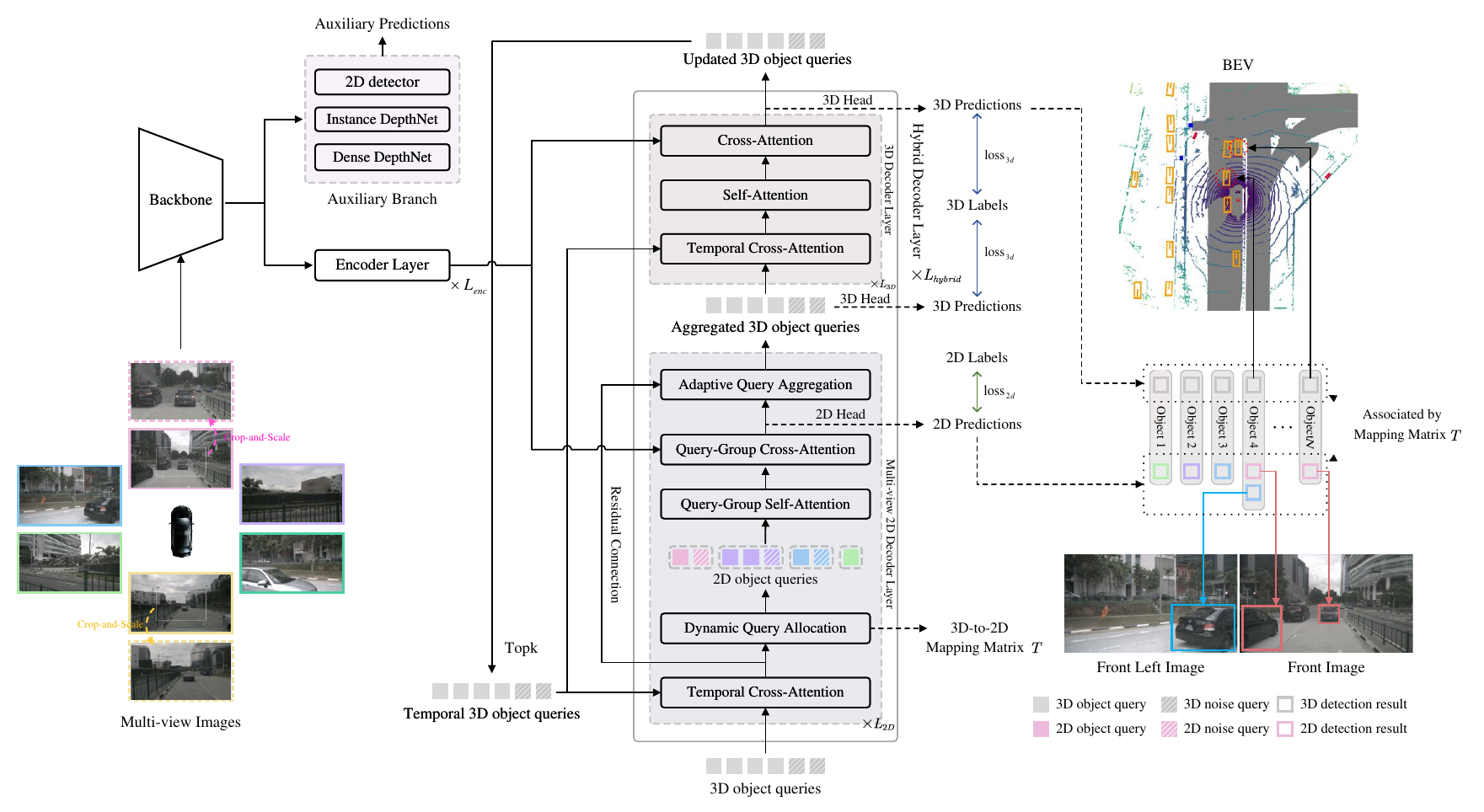}
\caption{Comprehensive architecture of SimPB++.}
\label{fig:sup_framework_pipline}
\end{figure*}

To maintain clarity, some minor components in Figure~\ref{fig:2_framework_pipeline} of the main manuscript have been omitted.
We provide the complete architecture of SimPB++ in Figure \ref{fig:sup_framework_pipline} for a comprehensive view.
Specifically, we include arrow lines to illustrate the connections between self-attention and cross-attention in both the multi-view 2D decoder layer and the 3D decoder layer. 
Additionally, we visualize the residual connection from the output of temporal cross-attention to the Adaptive Query Aggregation module. 
The aggregated 3D queries are separately shown as the output of a multi-view 2D decoder layer, which is used as input for the 3D head for deep supervision. 
Furthermore, to display the temporal propagation, we add an arrow line to indicate the updated object queries linking to the temporal 3D object queries.

In contrast to the preliminary architecture, SimPB++ extends the framework to support arbitrary multi-view input configurations. Based on this, we first implement a Crop-and-Scale strategy to construct a long-range view as shown in Figure \ref{fig:sup_framework_pipline} (left), the cropped and scaled region is highlighted by a dashed boundary. Secondly, we integrate auxiliary training branches, including a lightweight 2D detector and a DepthNet, targeting both dense maps and instance-level objects without increasing inference latency. Finally, to facilitate faster convergence, a query denoising strategy is applied to both 2D and 3D queries, which are represented as squares with slanted line in our diagrams.

\subsubsection{More Allocation Details}
In the Dynamic Query Allocation module, a 3D query is allocated to a maximum of one object center (anchor center) and multiple projection centers (box center) across different camera views by projecting it using camera parameters, as shown in Figure \ref{fig:3_allocation_module}. The projection center typically represents a truncated portion of a cross-view target. The total number of 2D object queries during the allocation process is equal to the combined count of object centers and projection centers.

During the early stages of training, the presence of inaccurate anchors can lead to a rapid increase in the number of projection centers, resulting in convergence challenges.
To address it, we introduce two constraint strategies to optimize the allocation during training.
\begin{itemize}
    \item The number of projection centers is limited to a maximum of $100$ for each camera group. Consequently, the total number of 2D queries is restricted to a maximum of $N + 100 \times V$, where $N$ represents the number of 3D queries and $V$ denotes the number of cameras.
    
    \item To mitigate the impact of incorrectly projected anchors, we limit the maximum size $\{l, w, h\}$ of the anchors to $\{35, 35, 10\}$, which is computed from the training split of nuScenes dataset.
\end{itemize}

For instance, while the number of 3D queries is fixed at $N=900$ across$V=6$ cameras, the number of 2D queries $M$ is dynamically adjusted based on anchor projections, averaging approximately 1,100. Despite this increased query count, the impact on computational overhead remains negligible.

\subsection{More Experimental Analysis}
\label{sec:experiment}

\subsubsection{About nuScenes test splits}
The groundtruth for the nuScenes test set has not been made publicly available. We are submitting our predictions to the EvalAI server for evaluation, as outlined in the official manual. However, prior to our paper submission, the EvalAI server is unable to provide test evaluation results, revealing stuck status. Many individuals have encountered the same issue, and it has been reported in the corresponding issues on the nuScenes and EvalAI official GitHub repositories, yet have not received any positive feedback regarding this matter.
As a result, the results of SimPB++ on the nuScenes test split cannot be presented in the paper. We will update the test results on our project website as soon as the evaluation available.

\subsubsection{Effect of Association}
\label{sec:effectiveness_association}

Two-stage methods usually only provide a default association between 3D and 2D results during query initialization (as shown in Figure \ref{fig:sup_difference_w_previous} (a)). In contrast, SimPB++ explicitly constructs the association of 2D and 3D detection results (as shown in Figure \ref{fig:sup_difference_w_previous} (c)). To quantitatively evaluate the association, we design a metric termed Association Accuracy Rate (AAR) to measure the accuracy rate of association and also apply Recall as an evaluation metric as well. 

Suppose there are $N_{\text{3d}}$ 3D groundtruth boxes $\{G_{3d}^{i}\}_{i=1}^{N_{\text{3d}}}$ and $N_{\text{2d}}$ projected 2D boxes $\{G_{2d}^{i}\}_{i=1}^{N_{\text{2d}}}$ on the image views as 2D boxes label in the validation dataset. We can obtain $M_{\text{3d}}$ 3D box prediction $\{P_{3d}^{i}\}_{i=1}^{M_{\text{3d}}}$ and $M_{\text{2d}}$ 2D prediction $\{P_{2d}^{i}\}_{i=1}^{M_{\text{2d}}}$ from the network.
Typically, a candidate match is established between a 3D prediction and a 2D groundtruth box. The total number of candidates matching is denoted as $\# \text{Matching}_\text{\{3D-pred, 2D-gt\}}$. From these candidate matches, we select valid associations between 3D predictions and 2D predictions generated by the network. The valid matching number is $\# \text{ValidMatching}_\text{\{3D-pred, 2D-pred\}}$. Therefore, we define the association evaluation metric AAR as follows:
\begin{gather}
    \text{AAR} =\cfrac{\# \text{ValidMatching}_\text{\{3D-pred, 2D-pred\}}}{\# \text{Matching}_\text{\{3D-pred, 2D-gt\}}} \times 100\%, \\
    \text{Recall} = \cfrac{\# \text{Matching}_\text{\{3D-pred, 2D-gt\}}}{N_{2d}} \times 100\%.
\end{gather} 

\begin{figure}[!t]
\centering
\includegraphics[width=\linewidth]{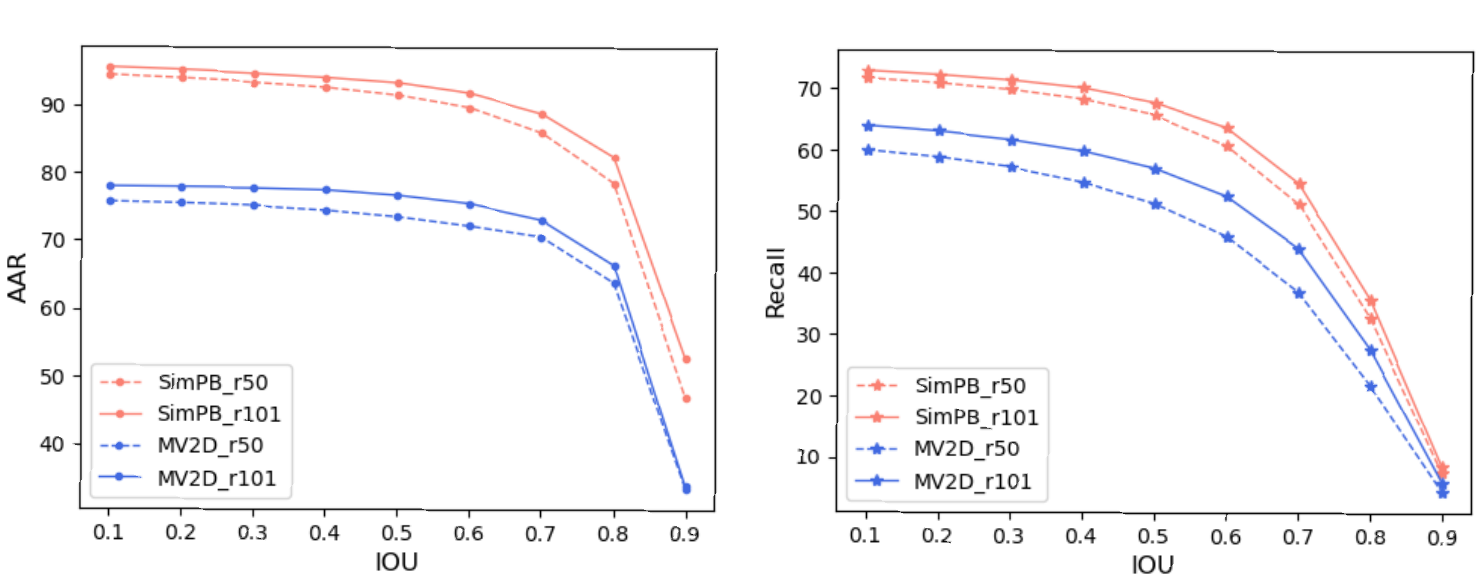}
\caption{AAR (Association Accuracy Rate) and Recall curves.}
\label{fig:sup_association_metric}
\end{figure}

For a given 3D prediction $P_{3d}^{i}$ and $j$-th 2D groundtruth $G_{2d}^{j}$ on $v$-th image view. And we denote the associated 3D groundtruth of $G_{2d}^{j}$ as $G_{3d}^{j}$ for simplicity. 
$P_{3d \rightarrow 2d}^{i}$ is the bounding rectangle on $v$-th view projected from $P_{3d}^{i}$.
The connection between $P_{3d}^{i}$ and $G_{2d}^{j}$ is a candidate matching if the following conditions are met:

\begin{equation}
\Phi(P_{3d}^{i}, G_{2d}^{j}) =
\begin{cases} 
    1 & \text{Dist}(P_{3d}^{i}, G_{3d}^{j}) \leq \tau_{dis} \ \& \\
      & \text{IoU}(P_{3d \rightarrow 2d}^{i}, G_{2d}^{j}) \geq \tau_{iou} \ \& \\
      & \text{Cls}(P_{3d}^{i}, G_{3d}^{j}) = 1 \\
    0 & \text{otherwise}
\end{cases},
\end{equation}
where $\text{Dist}$ represents the Euclidean distance between the centers of two 3D boxes, and $\text{Cls}$ indicates whether the labels of the two boxes are the same or not.
Similarly, we denote a valid candidate between the $i$-th 3D prediction $P_{3d}^{i}$ and the $k$-th 2D prediction $P_{2d}^{k}$ when the following conditions are met:

\begin{equation}
\Psi(P_{3d}^{i}, P_{2d}^{k}) =
\begin{cases} 
    1 & \Phi(P_{3d}^{i}, G_{2d}^{j})=1 \ \& \\ 
      & \text{IoU}(P_{2d}^{k} , G_{2d}^{j} ) \geq \tau_{iou}  \ \& \\
      & \text{Cls}(P_{2d}^{k}, G_{2d}^{j})=1 \\
    0 & \text{otherwise}
\end{cases}.
\end{equation}
Therefore, AAR can be rewritten as:
\begin{equation}
    \text{AAR}  =\cfrac{\sum_{i=1}^{M_{3d}}\sum_{k=1}^{M_{2d}} \Psi(P_{3d}^{i}, P_{2d}^{k}) } {\sum_{i=1}^{M_{3d}}\sum_{j=1}^{N_{2d}} \Phi(P_{3d}^{i}, G_{2d}^{j}) } \times 100\% .
\end{equation}

We fix the $\tau_{dis} = 2$ and adjust $\tau_{iou}$ from 0.1 to 0.9 to draw the AAR and Recall curves.
As shown in Figure \ref{fig:sup_association_metric}, the accuracy and recall decrease as the IOU threshold $\tau_{iou}$ increases. However, SimPB++ constantly gains higher AAR and Recall metrics with a large margin on both settings.
The utilization of the 3D-to-2D association in SimPB++ demonstrates higher accuracy compared to the 2D-to-3D association used in MV2D. This approach not only maintains a larger number of matched predictions but also ensures better alignment of 2D and 3D features. As a result, the 2D information to the same target is effectively leveraged for 3D tasks, leading to improved performance.

\subsection{More Visualization Results}
\label{sec:visualization}
We provide comprehensive visualizations in the supplementary material, including video sequences of SimPB++’s 2D and 3D detection performances on both nuScenes and Argoverse 2.

\end{document}